\documentclass{article}

\PassOptionsToPackage{numbers, compress}{natbib}
\usepackage[final]{neurips_2022}

\usepackage[utf8]{inputenc} 
\usepackage[T1]{fontenc}    
\usepackage[colorlinks, bookmarks=false]{hyperref}       
\usepackage{url}            
\usepackage{booktabs}       
\usepackage{amsfonts}       
\usepackage{nicefrac}       
\usepackage{microtype}      
\usepackage{xcolor}         

\usepackage{amsthm}
\usepackage{amsmath}
\usepackage{amssymb}
\usepackage{graphicx}
\usepackage{subfigure}
\usepackage{algorithm}
\usepackage{algorithmic}

\usepackage{fancybox}
\usepackage{multirow}
\usepackage{makecell}
\usepackage{xcolor}
\usepackage{caption}
\usepackage{multirow}
\usepackage[numbers]{natbib}
\usepackage{soul}
\usepackage{tablefootnote}

\usepackage{cleveref}
\crefname{figure}{Fig.}{Fig.}
\crefname{subfig}{Fig.}{Fig.}
\crefname{table}{Table}{Table}
\crefname{equation}{Equation}{Equation}
\crefname{section}{Section}{Section}
\crefname{algorithm}{Algorithm}{Algorithm}

\newtheorem{assumption}{Assumption}
\newtheorem{theorem}{Theorem}
\newtheorem{lemma}{Lemma}
\newtheorem{proposition}{Proposition}

\def \E {\mathbb{E}}
\def \w {\boldsymbol{w}}
\def \m {\boldsymbol{m}}
\def \x {\boldsymbol{x}}

\def \xi {\boldsymbol{x}_i}
\def \yi {\boldsymbol{y}_i}

\title{Make Sharpness-Aware Minimization Stronger: \\
A Sparsified Perturbation Approach}
 
\author{%
    Peng Mi$^1$\thanks{This work was done during an internship at JD Explore Academy.} \quad Li Shen$^2$\thanks{Li Shen is the corresponding author.} \quad Tianhe Ren$^1$ \quad Yiyi Zhou$^1$ \\ \quad \textbf{Xiaoshuai Sun}$^1$ \quad \textbf{Rongrong Ji}$^1$ \quad \textbf{Dacheng Tao}$^{2,3}$\\
    $^1$Media Analytics and Computing Laboratory, Department of Artificial Intelligence,\\
  School of Informatics, Xiamen University, China\\
    $^2$JD Explore Academy, Beijing, China \quad
    $^3$The University of Sydney, Australia\\
    \tt mipeng@stu.xmu.edu.cn, mathshenli@gmail.com, rentianhe@stu.xmu.edu.cn\\
    \tt zhouyiyi@xmu.edu.cn, xssun@xmu.edu.cn\\ 
    \tt rrji@xmu.edu.cn, dacheng.tao@gmail.com\\
}

\begin{document}

\maketitle

\begin{abstract}
Deep neural networks often suffer from poor generalization caused by complex and non-convex loss landscapes. One of the popular solutions is Sharpness-Aware Minimization (SAM), which smooths the loss landscape via minimizing the maximized change of training loss when adding a perturbation to the weight. However, we find the indiscriminate perturbation of SAM on all parameters is suboptimal, which also results in excessive computation,~\emph{i.e.}, double the overhead of common optimizers like Stochastic Gradient Descent~(SGD). In this paper, we propose an efficient and effective training scheme coined as Sparse SAM (SSAM), which achieves sparse perturbation by a binary mask. To obtain the sparse mask, we provide two solutions which are based onFisher information and dynamic sparse training, respectively. In addition, we theoretically prove that SSAM can converge at the same rate as SAM,~\emph{i.e.}, $O(\log T/\sqrt{T})$. Sparse SAM not only has the potential for training acceleration but also smooths the loss landscape effectively. Extensive experimental results on CIFAR10, CIFAR100, and ImageNet-1K confirm the superior efficiency of our method to SAM, and the performance is preserved or even better with a perturbation of merely 50\% sparsity. Code is available at \url{https://github.com/Mi-Peng/Sparse-Sharpness-Aware-Minimization}.
\end{abstract}
\section{Introduction}
\label[section]{sec:intro}
Over the past decade or so, the great success of deep learning has been due in great part to ever-larger model parameter sizes~\cite{vit, wideresnet, transformer, bert, swin-transformer, biggan}.
However, the excessive parameters also make the model inclined to poor generalization.
To overcome this problem, numerous efforts have been devoted to  training algorithm~\cite{early-stop, entropy-regularization, label-smoothing}, data augmentation~\cite{cutout, mixup, cutmix}, and network design~\cite{dropout, bn}. 

One important finding in recent research is the connection between the geometry of loss landscape and model generalization~\cite{LBtrain, sam, flat-minima, exploring-generalization,adversarial-robust}. In general, the loss landscape of the model is complex and non-convex, which makes model tend to converge to sharp minima. Recent endeavors~\cite{LBtrain, flat-minima, exploring-generalization} show that the flatter the minima of convergence, the better the model generalization. This discovery reveals the nature of previous approaches~\cite{early-stop, dropout, cutout, cutmix, mixup, bn} to improve generalization,~\emph{i.e.}, smoothing the loss landscape. 

Based on this finding, Foret~\emph{et al.}~\cite{sam} propose a novel approach to improve model generalization called \emph{sharpness-aware minimization} (SAM), which simultaneously minimizes loss value and loss sharpness. SAM quantifies the landscape sharpness as the maximized difference of loss when a perturbation is added to the weight. When the model reaches a sharp area, the perturbed gradients in SAM help the model jump out of the sharp minima. In practice, SAM requires two forward-backward computations for each optimization step, where the first computation is to obtain the perturbation and the second one is for parameter update. Despite the remarkable performance~\cite{sam,asam,esam,vit-sam}, This property makes SAM double the computational cost of the conventional optimizer,~\emph{e.g.}, SGD~\cite{bottou2010large}.

Since SAM calculates perturbations indiscriminately for all parameters, a question is arisen: 
\begin{center}
 \emph{Do we need to calculate perturbations for all parameters?}
\end{center}

Above all, we notice that in most deep neural networks, only about 5\% of parameters are sharp and rise steeply during optimization~\cite{LBtrain}. Then we explore the effect of SAM in different dimensions to answer the above question and find out \emph{(i) little difference between SGD and SAM gradients in most dimensions (see Fig.~\ref{fig:grad-diff}); (ii) more flatter without SAM in some dimensions (see Fig.~\ref{fig:vis-landscape} and Fig.~\ref{fig:hessian}).} 

Inspired by the above discoveries, we propose a novel scheme to improve the efficiency of SAM via sparse perturbation, termed Sparse SAM (SSAM). SSAM, which plays the role of regularization, has better generalization, and its sparse operation also ensures the efficiency of optimization.
Specifically, the perturbation in SSAM is multiplied by a binary sparse mask to determine which parameters should be perturbed. To obtain the sparse mask, we provide two implementations. The first solution is to use Fisher information~\cite{fisher-information} of the parameters to formulate the binary mask, dubbed SSAM-F. The other one is to employ dynamic sparse training to jointly optimize model parameters and the sparse mask, dubbed SSAM-D. The first solution is relatively more stable but a bit time-consuming, while the latter is more efficient.  

In addition to these solutions, we provide the theoretical convergence analysis of SAM and SSAM in non-convex stochastic setting, proving that our SSAM can converge at the same rate as SAM,~\emph{i.e.}, $O(\log T/\sqrt{T})$. At last, we evaluate the performance and effectiveness of SSAM on CIFAR10~\cite{cifar10/100}, CIFAR100~\cite{cifar10/100} and ImageNet~\cite{imagenet} with various models. The experiments confirm that SSAM contributes to a flatter landscape than SAM, and its performance is on par with or even better than SAM with only about 50\% perturbation. These results coincide with our motivations and findings.

To sum up, the contribution of this paper is three-fold:
\begin{itemize}
    \item We rethink the role of perturbation in SAM and find that the indiscriminate perturbations are suboptimal and computationally inefficient.
    \item We propose a sparsified perturbation approach called Sparse SAM (SSAM) with two variants,~\emph{i.e.}, Fisher SSAM (SSAM-F) and Dynamic SSAM (SSAM-D), both of which enjoy better efficiency and effectiveness than SAM. We also theoretically prove that SSAM can converge at the same rate as SAM,~\emph{i.e.}, $O(\log T/\sqrt{T})$. 
    \item We evaluate SSAM with various models on CIFAR and ImageNet, showing WideResNet with SSAM of a high sparsity outperforms SAM on CIFAR; SSAM can achieve competitive performance with a high sparsity; SSAM has a comparable convergence rate to SAM.
\end{itemize}

\section{Related Work}
\label[section]{sec:related-work}
In this section, we briefly review the studies on sharpness-aware minimum optimization (SAM), Fisher information in deep learning, and dynamic sparse training.

\textbf{SAM and flat minima.}
Hochreiter~\emph{et al.}~\cite{flat-minima} first reveal that there is a strong correlation between the generalization of a model and the flat minima. After that, there is a growing amount of research based on this finding. Keskar~\emph{et al.}~\cite{LBtrain} conduct experiments with a larger batch size, and in consequence observe the degradation of model generalization capability. They~\cite{LBtrain} also confirm the essence of this phenomenon, which is that the model tends to converge to the sharp minima. Keskar~\emph{et al.}~\cite{LBtrain} and Dinh~\emph{et al.}~\cite{sharp-can-generalize} state that the sharpness can be evaluated by the eigenvalues of the Hessian. However, they fail to find the flat minima due to the notorious computational cost of Hessian. 

Inspired by this, Foret~\emph{et al.}~\cite{sam} introduce a sharpness-aware optimization (SAM) to find a flat minimum for improving generalization capability, which is achieved by solving a mini-max problem. Zhang~\emph{et al.}~\cite{penalize-sam} make a point that SAM~\cite{sam} is equivalent to adding the regularization of the gradient norm by approximating Hessian matrix. Kwon~\emph{et al.}~\cite{asam} propose a scale-invariant SAM scheme with adaptive radius to improve training stability. Zhang~\emph{et al.}~\cite{surrogate} redefine the landscape sharpness from an intuitive and theoretical perspective based on SAM. To reduce the computational cost in SAM, Du~\emph{et al.}~\cite{esam} proposed Efficient SAM (ESAM) to randomly calculate perturbation. However, ESAM randomly select the samples every steps, which may lead to optimization bias. Instead of the perturbations for all parameters,~\emph{i.e.}, SAM, we compute a sparse perturbation,~\emph{i.e.}, SSAM, which learns important but sparse dimensions for perturbation.

\textbf{Fisher information (FI).} Fisher information
\cite{fisher-information} was proposed to measure the information that an observable random variable carries about an unknown parameter of a distribution.
In machine learning, Fisher information is widely used to measure the importance of the model parameters~\cite{pnas-forgetting-in-neural-networks} and decide which parameter to be pruned~\cite{fisher-mask, fisher-pruning}. 
For proposed SSAM-F, Fisher information is used to determine whether a weight should be perturbed for flat minima.  

\textbf{Dynamic sparse training.}
Finding the sparse network via pruning unimportant weights is a popular solution in network compression, which can be traced back to decades~\cite{lecun-1th-prune}. The widely used training scheme, \emph{i.e.}, pretraining-pruning-fine-tuning, is presented by Han~\emph{et.al.}~\cite{prune-pipeline}. Limited by the requirement for the pre-trained model, some recent research~\cite{rig,dst-recent-deep-rewiring,sparse-cosine,dst-recent-topkast,dst-recent-reparameterization,liu2021sparse,liu2022unreasonable} attempts to discover a sparse network directly from the training process. Dynamic Sparse Training (DST) finds the sparse structure by dynamic parameter reallocation. The criterion of pruning could be weight magnitude~\cite{LTH}, gradient~\cite{rig} and Hessian~\cite{lecun-1th-prune, woodfisher}, \emph{etc}.
We claim that different from the existing DST methods that prune neurons, our target is to obtain a binary mask for sparse perturbation. 
\section{Rethinking the Perturbation in SAM}
\label[section]{sec:rethinking}
In this section, we first review how SAM converges at the flat minimum of a loss landscape. Then, we rethink the role of perturbation in SAM.
 
\subsection{Preliminary}
\label[section]{sec:rethinking:preliminary}

In this paper, we consider the weights of a deep neural network as $\w=(w_1,w_2,...,w_d) \subseteq \mathcal{W}\in \mathbb{R}^d$ and denote a binary mask as $\m \in \{0,1\}^d$, which satisfies $\boldsymbol{1}^T\m=(1-s)\cdot d$ to restrict the computational cost. Given a training dataset as $\mathcal{S} \triangleq \{(\xi,\yi)\}_{i=1}^n$ \emph{i.i.d.} drawn from the distribution $\mathcal{D}$, the per-data-point loss function is defined by $f(\w, \xi, \yi)$. For the classification task in this paper, we use cross-entropy as loss function. The population loss is defined by $f_{\mathcal{D}}=\E_{(\xi,\yi)\sim\mathcal{D}}f(\w,\xi,\yi)$, while the empirical training loss function is $f_{\mathcal{S}}\triangleq \frac{1}{n}\sum_{i=1}^nf(\w,\xi,\yi)$.

Sharpness-aware minimization (SAM)~\cite{sam} aims to simultaneously minimize the loss value and smooth the loss landscape, which is achieved by solving the min-max problem:
\begin{equation}
\label[equation]{equ:sam}
    \min_{\w} \max_{||\boldsymbol{\epsilon}||_2\leq\rho} f_{\mathcal{S}}(\w + \boldsymbol{\epsilon}).
\end{equation}
SAM first obtains the perturbation $\boldsymbol{\epsilon}$ in a neighborhood ball area with a radius denoted as $\rho$. The optimization tries to minimize the loss of the perturbed weight $\w + \boldsymbol{\epsilon}$. Intuitively, the goal of SAM is that small perturbations to the weight will not significantly rise the empirical loss, which indicates that SAM tends to converge to a flat minimum. To solve the mini-max optimization, SAM approximately calculates the perturbations $\boldsymbol{\epsilon}$ using Taylor expansion around $\w$:
\begin{equation}
\label[equation]{equ:eps}
\boldsymbol{\epsilon} = \mathop{\arg\max}_{||\boldsymbol{\epsilon}||_2\leq\rho} f_{\mathcal{S}}(\w + \boldsymbol{\epsilon}) \approx \mathop{\arg\max}_{||\boldsymbol{\epsilon}||_2\leq\rho} f_{\mathcal{S}}(\w) + \boldsymbol{\epsilon} \cdot \nabla_{\w}f(\w) = \rho \cdot {\nabla_{\w}f(\w)}{\big /}{||\nabla_{\w}f(\w)||_2}.
\end{equation}
In this way, the objective function can be rewritten as $\min_{\w}f_{\mathcal{S}}\big(\w+\rho {\nabla_{\w}f(\w)}{\big/}{||\nabla_{\w}f(\w)||_2}\big)$, which could be implemented by a two-step gradient descent framework in Pytorch or TensorFlow:
\begin{itemize}
    \item In the first step, the gradient at $\w$ is used to calculate the perturbation $\boldsymbol{\epsilon}$ by Eq. \ref{equ:eps}. Then the weight of model will be added to $\w + \boldsymbol{\epsilon}$.
    \item In the second step, the gradient at $\w + \boldsymbol{\epsilon}$ is used to solve $\min_{\w} f_{\mathcal{S}}(w+\boldsymbol{\epsilon})$, \emph{i.e.}, update the weight $\w$ by this gradient.
\end{itemize}

\subsection{Rethinking the Perturbation Step of SAM}
\label[section]{sec:rethinking:rethinkSAM}

\textbf{How does SAM work in flat subspace?} SAM perturbs all parameters indiscriminately, but the fact is that merely about 5\% parameter space is sharp while the rest is flat~\cite{LBtrain}. We are curious whether perturbing the parameters in those already flat dimensions would lead to the instability of the optimization and impair the improvement of generalization. To answer this question, we quantitatively and qualitatively analyze the loss landscapes with different training schemes in~\cref{sec:experiments}, as shown in ~\cref{fig:vis-landscape} and~\cref{fig:hessian}. The results confirm our conjecture that optimizing some dimensions without perturbation can help the model generalize better. 

\begin{figure}[h]
    \centering
    \vspace{-0.2cm}
    \includegraphics[width=0.98\linewidth]{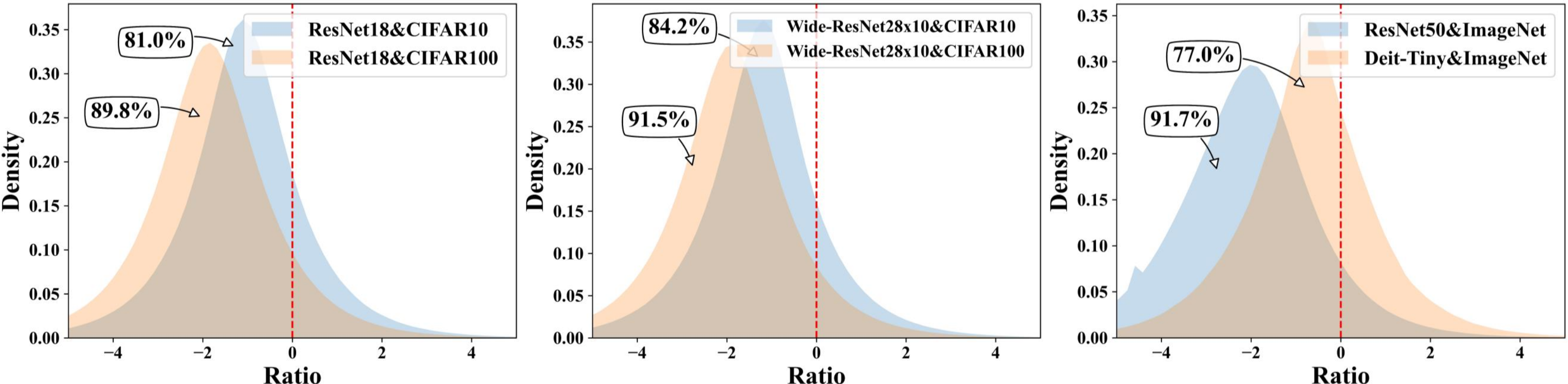}
    \vspace{-0.1cm}
    \caption{The distribution of relative difference ratio $r$ among various models and datasets. There is little difference between SAM and SGD gradients for most parameters,~\emph{i.e.}, the ratio $r$ is less than $0$.}
    \label[figure]{fig:grad-diff}
    \vspace{-0.4cm}
\end{figure}

\textbf{What is the difference between the gradients of SGD and SAM?}\ 
We investigate various neural networks optimized with SAM and SGD on CIFAR10/100 and ImageNet, whose statistics are given in~\cref{fig:grad-diff}. We use the relative difference ratio $r$, defined as $r = \log \left|{(g_{SAM}- g_{SGD})}/{g_{SGD}}\right|$, to measure the difference between the gradients of SAM and SGD. As showin in~\cref{fig:grad-diff}, the parameters with $r$ less than 0 account for the vast majority of all parameters, indicating that most SAM gradients are not significantly different from SGD gradients. These results show that most parameters of the model require no perturbations for achieving the flat minima, which well confirms our the motivation. 

Inspired by the above observation and the promising hardware acceleration for sparse operation modern GPUs, we further propose Sparse SAM, a novel sparse perturbation approach, as an implicit regularization to improve the efficiency and effectiveness of SAM.
\section{Methodology}
\label[section]{sec:method}
In this section, we first define the proposed Sparse SAM (SSAM), which strengths SAM via sparse perturbation. Afterwards, we introduce the instantiations of the sparse mask used in SSAM via Fisher information and dynamic sparse training, dubbed SSAM-F and SSAM-D, respectively. 

\subsection{Sparse SAM}
\label[section]{sec:method:sparseSAM}
Motivated by the finding discussed in the introduction, Sparse SAM (SSAM) employs a sparse binary mask to decide which parameters should be perturbed, thereby improving the efficiency of sharpness-aware minimization.  
Specifically, the perturbation $\boldsymbol{\epsilon}$ will be multiplied by a sparse binary mask $\m$, and the objective function is then rewritten as $\min_{\w}f_{\mathcal{S}}\left(\w+\rho\cdot\frac{\nabla_{\w} f(\w)}{||\nabla_{\w}f(\w)||_2}\odot \m\right)$.  To stable the optimization, the sparse binary mask $\m$ is updated at regular intervals during training. We provide two solutions to obtain the sparse mask $\m$, namely Fisher information based Sparse SAM (SSAM-F) and dynamic sparse training based Sparse SAM (SSAM-D). The overall algorithms of SSAM and sparse mask generations are shown in Algorithm~\ref{alg:sparse-sam-pipeline} and Algorithm~\ref{alg:sparse-sam-generate-mask}, respectively.

\begin{table}[ht]
\centering
\vspace{-0.6cm}
    \begin{tabular}{c c}
    \hspace{-0.8cm}
\begin{minipage}[t]{0.44\linewidth}
    \vspace{0pt}
    \begin{algorithm}[H]
    \small
        \caption{Sparse SAM (SSAM)}
        \label{alg:sparse-sam-pipeline}
    \begin{algorithmic}[1]
    \REQUIRE sparse ratio $s$, dense model $\w$, binary mask $\m$, update interval $T_m$, number of samples $N_F$, learning rate $\eta$, training set $\mathcal{S}$.
    \STATE Initialize $\w$ and $\m$ randomly.
    \FOR{epoch $t=1,2 \ldots T$}
        \FOR{each training iteration}
        \STATE{Sample a batch from $\mathcal{S}$: $\mathcal{B}$}
        \STATE{Compute perturbation $\boldsymbol{\epsilon}$ by Eq.~\eqref{equ:eps}}
        \IF{$t \mod T_m = 0$}
        \STATE Generate mask $m$ via {\color{blue}{Option I}} or {\color{blue}{II}}
        \ENDIF
        \STATE{$\boldsymbol{\epsilon}\gets\boldsymbol{\epsilon} \odot \m$}
        \ENDFOR
    \STATE{$\w \gets \w - \eta \nabla f(\w+\boldsymbol{\epsilon})$}
    \ENDFOR
    \RETURN{Final weight of model $\w$}
    \end{algorithmic}
    \end{algorithm}
\end{minipage}
\hspace{-2mm}
&
\begin{minipage}[t]{0.55\linewidth}
    \vspace{0pt}
    \begin{algorithm}[H]
    \small
         \caption{Sparse Mask Generation}
        \label{alg:sparse-sam-generate-mask}
    \begin{algorithmic}[1]
    \STATE  {\color{blue}{Option I:(Fisher Information Mask)}}
        \STATE{Sample $N_F$ data from $\mathcal{S}$: $\mathcal{B}_F$}
        \STATE{Compute Fisher $\hat{F}_{\w}$ by Eq.~\eqref{equ:fisher-empirical}}
        \STATE{$\boldsymbol{m_1}\!=\!\{m_i\!=\!1|m_i\!\in\! \m\}\gets {\rm ArgTopK}(\hat{F}_{\w}, s\cdot|\w|)$}
        \STATE{$\boldsymbol{m_0}\!=\!\{m_i\!=\!0|m_i\!\in\! \m\}\gets \{m_i|m_i\!\notin\!\boldsymbol{m_1}\}$}
        \STATE{$\m \gets \boldsymbol{m_0} \cup \boldsymbol{m_1}$}
    \STATE  {\color{blue}{Option II:(Dynamic Sparse Mask)}}    
        \STATE{$N_{drop}=f_{decay}(t;\alpha)\cdot (1 - s) \cdot |\w|$}
        \STATE{$N_{growth}=N_{drop}$}
        \STATE{$\boldsymbol{m_1}\!=\!\{m_i\!=\!1|m_i\!\in\! \m\}\!\gets\!\{m_i\!=\!1|m_i\!\in\!\m\}\!-\! {\rm ArgTopK}_{m_i\!\in\!\boldsymbol{m_1}}(-|\nabla f(\w)|, N_{drop})$}
        \STATE{$\boldsymbol{m_1}\!\gets\!\{m_i\!=\!1|m_i\!\in\!\m\} +{\rm Random}_{\m_i\!\notin\!\boldsymbol{m_1}}(N_{growth})$}
        \STATE{$\boldsymbol{m_0}\!=\!\{m_i\!=\!0|m_i\!\in\! \m\}\gets \{m_i|m_i\!\notin\!\boldsymbol{m_1}\}$}
        \STATE{$\m \gets \boldsymbol{m_0} \cup \boldsymbol{m_1}$}
    \RETURN Sparse mask $\m$
    \end{algorithmic}
    \end{algorithm}
\end{minipage}
    \end{tabular}
\end{table}

According to the previous work~\cite{2:4-truly-sparse} and Ampere architecture equipped with sparse tensor cores~\cite{nvidia-whitepaper, nvidia-asp, nvidia-ampere-tensor-core}, currently there exists technical support for matrix multiplication with 50\% fine-grained sparsity~\cite{2:4-truly-sparse}\footnote{For instance, 2:4 sparsity for A100 GPU.}. Therefore, SSAM of 50\% sparse perturbation has great potential to achieve true training acceleration via sparse back-propagation.

\subsection{SSAM-F: Fisher information based Sparse SAM}
\label[section]{sec:method:sparseSAM:fisherSAM}

Inspired by the connection between Fisher information and Hessian~\cite{fisher-information}, which can directly measure the flatness of the loss landscape, we apply Fisher information to achieve sparse perturbation, denoted as SSAM-F. The Fisher information is defined by
\begin{equation}
\label[Equation]{equ:fisher-information-matrix}
    F_{\w} = \E_{x\sim p(x)}\left[ \E_{y\sim p_{\w}(y|x)}\nabla_{\w}\log p_{\w}(y|x) \nabla_{\w}\log p_{\w}(y|x)^T\right],
\end{equation} 
where $p_{\w}(y|x)$ is the output distribution predicted by the model. In over-parameterized networks, the computation of Fisher information matrix is also intractable, \emph{i.e.}, $|\w|\times|\w|$. Following~\cite{fisher-mask, KFAC, KFAC-opt, KFAC-cnn}, we approximate $F_{\w}$ as a diagonal matrix, which is included in a vector in $\mathbb{R}^{|\w|}$. Note that there are the two expectation in Eq.~\eqref{equ:fisher-information-matrix}. The first one is that the original data distribution $p(x)$ is often not available. Therefore, we approximate it by sampling $N_F$ data $\x_1,\x_2,\ldots, \x_{N_F}$ from $x\sim p(x)$:
\begin{equation}
\label[equation]{equ:fisher-approximation-x}
    F_{\w}=\frac{1}{N_F}\E_{y\sim p_{\w}(y|\x_i)}(\nabla_{\w}\log p_{\w}(y|\x_i))^2.
\end{equation}
For the second expectation over $p_{\w}(y|\x)$, it is not necessary to compute its explicit expectation, since the ground-truth $y_i$ for each training  sample $x_i$ is available in supervised learning. Therefore we rewrite the Eq.~\eqref{equ:fisher-approximation-x} as  "empirical Fisher":
\begin{equation}
\label[equation]{equ:fisher-empirical}
\hat{F}_{\w}=\frac{1}{N_F}(\nabla_{\w}\log p_{\w}(y_i|\x_i))^2.
\end{equation}
We emphasize that the empirical Fisher is a $|\w|$-dimension vector, which is the same as the mask $\m$. To obtain the mask $\m$, we calculate the empirical Fisher by Eq.~\eqref{equ:fisher-empirical} over $N_F$ training data randomly sampled from training set $\mathcal{S}$. Then we sort the elements of empirical Fisher in descending, and the parameters corresponding to the the top $k$ Fisher values will be perturbed:
\begin{equation}
\boldsymbol{m_1}\!=\!\{m_i\!=\!1|m_i\!\in\! \m\}\gets {\rm ArgTopK}(\hat{F}_{\w}, k),
\end{equation}
where ${\rm ArgTopK}(\boldsymbol{v}, N)$ returns the index of the top $N$ largest values among $\boldsymbol{v}$, $\boldsymbol{m_1}$ is the set of values that are 1 in $\m$. $k$ is the number of perturbed parameters, which is equal to $(1-s)\cdot |\w|$ for sparsity $s$. After setting the rest values of the mask to 0,~\emph{i.e.}, $\boldsymbol{m_0}=\{m_i=0|m_i\notin \boldsymbol{m_1}\}$, we get the final mask $\m=\boldsymbol{m_0}\cup\boldsymbol{m_1}$. The algorithm of SSAM-F is shown in \cref{alg:sparse-sam-generate-mask}.

\subsection{SSAM-D: Dynamic sparse training mask based sparse SAM}
\label[section]{sec:method:sparseSAM:dstSAM}
Considering the computation of empirical Fisher is still relatively high, we also resort to dynamic sparse training for efficient binary mask generation. The mask generation includes the perturbation dropping and the perturbation growth steps. At the perturbation dropping phase, the flattest dimensions of the perturbed parameters will be dropped,~\emph{i.e.}, the gradients of lower absolute values, which means that they require no perturbations. The update of the sparse mask follows
\begin{equation}
    \boldsymbol{m_1} = \{m_i=1|m_i\in\m\} \gets \boldsymbol{m_1} - {\rm ArgTopK}_{\w \in \boldsymbol{m_1}}(-|\nabla f(\w)|, N_{drop}),
\end{equation}
where $N_{drop}$ is the number of perturbations to be dropped. At the perturbation growth phase, for the purpose of exploring the perturbation combinations as many as possible, several unperturbed dimensions grow, which means these dimensions need to compute perturbations. The update of the sparse mask for perturbation growth follows
\begin{equation}
    \boldsymbol{m_1} = \{m_i=1|m_i\in\m\} \gets \boldsymbol{m_1} + {\rm Random}_{\w \notin \boldsymbol{m_1}}(N_{growth}),
\end{equation}
where ${\rm Random}_{\mathcal{S}}(N)$ randomly returns $N$ indexes in $\mathcal{S}$, and $N_{growth}$ is the number of perturbation growths. To keep the sparsity constant during training, the number of growths is equal to the number of dropping,~\emph{i.e.}, $N_{growth}=N_{drop}$. Afterwards, we set the rest values of the mask to $0$,~\emph{i.e.}, $\boldsymbol{m_0}\!=\!\{m_i\!=\!0|m_i\!\notin\!\boldsymbol{m_1}\}$, and get the final mask $\m\!=\!\boldsymbol{m_0}\cup\boldsymbol{m_1}$. The drop ratio $\alpha$ represents the proportion of dropped perturbations in the total perturbations $s\cdot|\w|$,~\emph{i.e.}, $\alpha=N_{drop}/(s\cdot|\w|)$. In particular, a larger drop rate means that more combinations of binary mask can be explored during optimization, which, however, may slightly interfere the optimization process. Following \cite{rig, sparse-cosine}, we apply a cosine decay scheduler to alleviate this problem:
\begin{equation}
    f_{decay}(t;\alpha) = \frac{\alpha}{2}\left( 1+\cos \left( {t\pi}{/}{T} \right)  \right),
\end{equation}
where $T$ denotes number of training epochs. The algorithm of SSAM-D is depicted in ~\cref{alg:sparse-sam-generate-mask}.

\subsection{Theoretical analysis of Sparse SAM}
\label[section]{sec:method:theoretical}
In the following, we analyze the convergence of SAM and SSAM in non-convex stochastic setting. Before introducing the main theorem, we first describe the following assumptions that are commonly used for characterizing the convergence of nonconvex stochastic optimization \cite{con-adv, surrogate, understand-sam,chen2022towards,bottou2018optimization,ghadimi2013stochastic}.
\begin{assumption}
\label{assume:bounded-gradient}
(Bounded Gradient.) It exists $G \geq 0$ s.t. $||\nabla f(\w)|| \leq G$.
\end{assumption}

\begin{assumption}
\label{assume:bounded-variance}
(Bounded Variance.) It exists $\sigma \geq 0$ s.t. $\E [||g(\w) - \nabla f(\w)||^2] \leq \sigma^2$.
\end{assumption}

\begin{assumption}
\label{assume:l-smoothness}
(L-smoothness.) It exists $L > 0 $ s.t. $||\nabla f(\w) - \nabla f(\boldsymbol{v})|| \leq L||\w - \boldsymbol{v}|| $, $\forall  \w, \boldsymbol{v} \in \mathbb{R}^d$.
\end{assumption}

\begin{theorem}
\label{theorem:sam}
Consider function $f(\w)$ satisfying the Assumptions
\ref{assume:bounded-gradient}-\ref{assume:l-smoothness} optimized by SAM. Let $\eta_t = \frac{\eta_0}{\sqrt{t}}$ and perturbation amplitude $\rho$ decay with square root of $t$, \emph{e.g.}, $\rho_t=\frac{\rho_0}{\sqrt{t}}$. With $\rho_0 \leq G \eta_0$, we have
\begin{equation}
    \frac{1}{T} \sum_{t=0}^T \E ||\nabla f(\w_t)||^2 \leq C_1 \frac{1}{\sqrt{T}} + C_2\frac{\log T}{\sqrt{T}},
\end{equation}
where $C_1=\frac{2}{\eta_0}(f(\w_0)-\E f(\w_T))$ and $C_2=2(L \sigma^2 \eta_0 + LG\rho_0)$. 
\end{theorem}

\begin{theorem}
\label{theorem:ssam}
Consider function $f(\w)$ satisfying the Assumptions
\ref{assume:bounded-gradient}-\ref{assume:l-smoothness} optimized by SSAM. Let $\eta_t=\frac{\eta_0}{\sqrt{t}}$ and perturbation amplitude $\rho$ decay with square root of $t$, \emph{e.g.}, $\rho_t=\frac{\rho_0}{\sqrt{t}}$. With $\rho_0 \leq G\eta_0/2$, we have:
\begin{equation}
    \frac{1}{T}\sum_{t=0}^T \E||\nabla f(\w_t)||^2 \leq
    C_3\frac{1}{\sqrt{T}} + C_4\frac{\log T}{\sqrt{T}},
\end{equation}
where $C_3=\frac{2}{\eta_0}(f(\w_0)-\E f(\w_T)+\eta_0L^2\rho^2(1+\eta_0L)\frac{\pi^2}{6})$ and $C_4=2(L\sigma^2\eta_0+LG\rho_0)$.
\end{theorem}

For non-convex stochastic optimization, Theorems \ref{theorem:sam}\&\ref{theorem:ssam} imply that our SSAM could converge the same rate as SAM,~\emph{i.e.}, $O(\log T/\sqrt{T})$. Detailed proofs of the two theorems are given in \textbf{Appendix}.
\section{Experiments}
\label[section]{sec:experiments}

In this section, we evaluate the effectiveness of  SSAM through extensive experiments on CIFAR10, CIFAR100~\cite{cifar10/100} and ImageNet-1K~\cite{imagenet}. The base models  include ResNet~\cite{resnet} and WideResNet~\cite{wideresnet}. 
We report the main results on CIFAR datasets in Tables \ref{resnet18-cifar-experiments}\&\ref{wideresnet-cifar-experiments} and ImageNet-1K  in~\cref{imagenet-experiments}. Then, we visualize the landscapes and Hessian spectra to verify that the proposed SSAM can help the model generalize better. More experimental results are placed in \textbf{Appendix} due to page limit.

\subsection{Implementation details}
\label{sec:experiments-implement-detail}
\textbf{Datasets.} 
We use CIFAR10/CIFAR100~\cite{cifar10/100} and ImageNet-1K~\cite{imagenet} as the benchmarks of our method. Specifically, CIFAR10 and CIFAR100 have 50,000 images of 32$\times$32 resolution for training, while 10,000 images for test. ImageNet-1K~\cite{imagenet} is the most widely used benchmark for image classification, which has 1,281,167 images of 1000 classes and 50,000 images for validation. 

\textbf{Hyper-parameter setting.} 
For small resolution datasets,~\emph{i.e.}, CIFAR10 and CIFAR100, we replace the first convolution layer in ResNet and WideResNet with the one of 3$\times$3 kernel size, 1 stride and 1 padding. The models on CIFAR10/CIFAR100 are trained with 128 batch size for 200 epochs. We apply the random crop, random horizontal flip, normalization and cutout~\cite{cutout} for data augmentation, and the initial learning rate is 0.05 with a cosine learning rate schedule. The momentum and weight decay of SGD are set to 0.9 and 5\emph{e}-4, respectively. SAM and SSAM apply the same settings, except that weight decay is set to 0.001 \cite{esam}. We determine the perturbation magnitude $\rho$ from $\{0.01, 0.02, 0.05, 0.1, 0.2, 0.5\}$ via grid search. In CIFAR10 and CIFAR100, we set $\rho$ as 0.1 and 0.2, respectively. 
For ImageNet-1K, we randomly resize and crop all images to a resolution of 224$\times$224, and apply random horizontal flip, normalization  during training. We train ResNet with a batch size of 256, and adopt the cosine learning rate schedule with initial learning rate 0.1.  The momentum and weight decay of SGD is set as 0.9 and 1\emph{e}-4. SAM and SSAM use the same settings as above. The test images of both architectures are resized to 256$\times$256 and then centerly cropped to 224$\times$224. The perturbation magnitude $\rho$ is set to 0.07.

\subsection{Experimental results}
\label[section]{sec:experiments:results}

\textbf{Results on CIFAR10/CIFAR100.}\
We first evaluate our SSAM on CIFAR-10 and CIFAR100. The models we used are ResNet-18~\cite{resnet} and WideResNet-28-10~\cite{wideresnet}. 
The perturbation magnitude $\rho$ for SAM and SSAM are the same for a fair comparison. As shown in~\cref{resnet18-cifar-experiments}, ResNet18 with SSAM of 50\% sparsity outperforms SAM of full perturbations. From~\cref{wideresnet-cifar-experiments}, we can observe that the advantages of SSAM-F and SSAM-D on WideResNet28 are more significant, which achieve better performance than SAM with up to 95\% sparsity. Note that the parameter size of WideResNet28 is much larger than that of ResNet-18, and CIFAR is often easy to overfit. In addition, even with very large sparsity, both SSAM-F and SSAM-D can still obtain competitive performance against SAM. 

\begin{table}[ht]
\vspace{-0.4cm}
\small
\caption{Comparison between SGD, SAM and SSAM on CIFAR using ResNet-18}
\vspace{0.1cm}
\label[table]{resnet18-cifar-experiments}
\centering
\begin{tabular}{cccccc}
\toprule
Model & Optimizer& Sparsity & CIFAR10 & CIFAR100 & FLOPs\tablefootnote{The FLOPs is theoretically estimated by considering sparse multiplication compared to SGD in Tables \ref{resnet18-cifar-experiments}- \ref{imagenet-experiments}.}
\\ \hline
\multirow{14}{*}{ResNet18} & SGD & / & 96.07\% & 77.80\% & 1$\times$ \\
& SAM & 0\% & 96.83\% & 81.03\% & 2$\times$\\ \cline{2-6} 
& \multirow{6}{*}{SSAM-F (Ours)} & 50\% & \textbf{96.81\%} \textcolor{blue}{(-0.02)} &  \textbf{81.24\%} \textcolor{blue}{(+0.21)} & 1.65$\times$ \\
& & 80\% & 96.64\% \textcolor{blue}{(-0.19)} & 80.47\%  \textcolor{blue}{(-0.56)} & 1.44$\times$ \\
& & 90\% & 96.75\% \textcolor{blue}{(-0.08)} & 80.02\%  \textcolor{blue}{(-1.01)} & 1.36$\times$ \\
& & 95\% & 96.66\% \textcolor{blue}{(-0.17)} & 80.50\%  \textcolor{blue}{(-0.53)} & 1.33$\times$ \\
& & 98\% & 96.55\% \textcolor{blue}{(-0.28)} & 80.09\%  \textcolor{blue}{(-0.94)} & 1.31$\times$ \\
& & 99\% & 96.52\% \textcolor{blue}{(-0.31)} & 80.07\% \textcolor{blue}{(-0.96)} & 1.30$\times$ \\
\cline{2-6}
& \multirow{6}{*}{SSAM-D (Ours)} & 50\% & \textbf{96.87\%} \textcolor{blue}{(+0.04)} & \textbf{80.59\%} \textcolor{blue}{(-0.44)} & 1.65$\times$ \\
& & 80\% & 96.76\% \textcolor{blue}{(-0.07)} & 80.43\% \textcolor{blue}{(-0.60)} & 1.44$\times$ \\
& & 90\% & 96.67\% \textcolor{blue}{(-0.16)} & 80.39\% \textcolor{blue}{(-0.64)} & 1.36$\times$ \\
& & 95\% & 96.56\% \textcolor{blue}{(-0.27)} & 79.79\% \textcolor{blue}{(-1.24)} & 1.33$\times$ \\
& & 98\% & 96.61\% \textcolor{blue}{(-0.22)} & 79.79\% \textcolor{blue}{(-1.24)} & 1.31$\times$ \\
& & 99\% & 96.59\% \textcolor{blue}{(-0.24)} & 79.61\% \textcolor{blue}{(-1.42)} & 1.30$\times$ \\ \bottomrule
\end{tabular}
\end{table}

\begin{table}[h]
\small
\caption{Comparsion between SGD, SAM and SSAM on CIFAR using WideResNet-28-10}
\vspace{0.1cm}
\label[table]{wideresnet-cifar-experiments}
\centering
\begin{tabular}{cccccc}
\toprule
Model & Optimizer & Sparsity & CIFAR10 & CIFAR100 & FLOPs\\ \hline
\multirow{14}{*}{\rotatebox{90}{WideResNet28-10}} & SGD & / & 97.11\% & 81.93\% & 1$\times$ \\
& SAM & 0\% & 97.48\% & 84.20\% & 2$\times$ \\ \cline{2-6} 
& \multirow{6}{*}{SSAM-F (Ours)} & 50\% & \textbf{97.71\%} \textcolor{blue}{(+0.23)} & \textbf{85.16\%} \textcolor{blue}{(+0.96)} & 1.65$\times$ \\
& & 80\% & 97.67\% \textcolor{blue}{(+0.19)} & 84.57\% \textcolor{blue}{(+0.37)} & 1.44$\times$ \\
& & 90\% & 97.47\% \textcolor{blue}{(-0.01)} & 84.76\% \textcolor{blue}{(+0.56)} & 1.36$\times$ \\
& & 95\% & 97.42\% \textcolor{blue}{(-0.06)} & 84.17\% \textcolor{blue}{(-0.03)} & 1.33$\times$ \\
& & 98\% & 97.32\% \textcolor{blue}{(-0.16)} & 83.85\% \textcolor{blue}{(-0.35)} & 1.31$\times$ \\
& & 99\% & 97.59\% \textcolor{blue}{(+0.11)} & 84.00\% \textcolor{blue}{(-0.20)} & 1.30$\times$ \\
\cline{2-6} 
& \multirow{6}{*}{SSAM-D (Ours)} & 50\% &  97.70\% \textcolor{blue}{(+0.22)}&  \textbf{84.99\%} \textcolor{blue}{(+0.79)} & 1.65$\times$ \\
& & 80\% & \textbf{97.72\%} \textcolor{blue}{(+0.24)} & 84.36\%  \textcolor{blue}{(+0.16)} & 1.44$\times$ \\
& & 90\% & 97.53\% \textcolor{blue}{(+0.05)} & 84.16\%  \textcolor{blue}{(-0.04)} & 1.36$\times$ \\
& & 95\% & 97.52\% \textcolor{blue}{(+0.04)} & 83.66\%  \textcolor{blue}{(-0.54)} & 1.33$\times$ \\
& & 98\% & 97.30\% \textcolor{blue}{(-0.18)} & 83.30\%  \textcolor{blue}{(-0.90)} & 1.31$\times$ \\
& & 99\% & 97.27\% \textcolor{blue}{(-0.25)} & 84.16\%  \textcolor{blue}{(-0.04)} & 1.30$\times$ \\ \bottomrule
\end{tabular}
\vspace{-0.4cm}
\end{table}

\textbf{Results on ImageNet.}\ 
~\cref{imagenet-experiments} reports the result of SSAM-F and SSAM-D on the large-scale ImageNet-1K~\cite{imagenet} dataset. The model we used is ResNet50~\cite{resnet}. We can observe that SSAM-F and SSAM-D can maintain the performance with 50\% sparsity. However, as sparsity ratio increases, SSAM-F will receive relatively obvious performance drops while SSAM-D is more robust. 

\begin{table}[ht]
\small
\caption{Comparsion between SGD, SAM and SSAM on ImageNet-1K using ResNet50.}
\label[table]{imagenet-experiments}
\centering
\begin{tabular}{ccccc}
\toprule
Model & Optimizer & Sparsity & ImageNet & FLOPs \\ \hline
\multirow{8}{*}{ResNet50} & SGD  & /  & 76.67\% & 1$\times$ \\ 
& SAM & 0\% & 77.25\% & 2$\times$ \\ \cline{2-5} 
& \multirow{3}{*}{SSAM-F (Ours)} & 50\% & \textbf{77.31\%}\textcolor{blue}{(+0.06)} & 1.65$\times$ \\
& & 80\% & 76.81\%\textcolor{blue}{(-0.44)} & 1.44$\times$ \\
& & 90\% & 76.74\%\textcolor{blue}{(-0.51)} & 1.36$\times$ \\ \cline{2-5} 
& \multirow{3}{*}{SSAM-D (Ours)} & 50\% & \textbf{77.25\%}\textcolor{blue}{(-0.00)} & 1.65$\times$ \\
& & 80\% & 77.00\%\textcolor{blue}{(-0.25)} & 1.44$\times$ \\
& & 90\% & 77.00\%\textcolor{blue}{(-0.25)} & 1.36$\times$ \\  \bottomrule
\end{tabular}
\vspace{-0.2cm}
\end{table}

\begin{figure}[H]
\vspace{-0.4cm}
\centering
\subfigure[ResNet18 on CIFAR10]{
    \centering
    \includegraphics[width=0.32\linewidth]{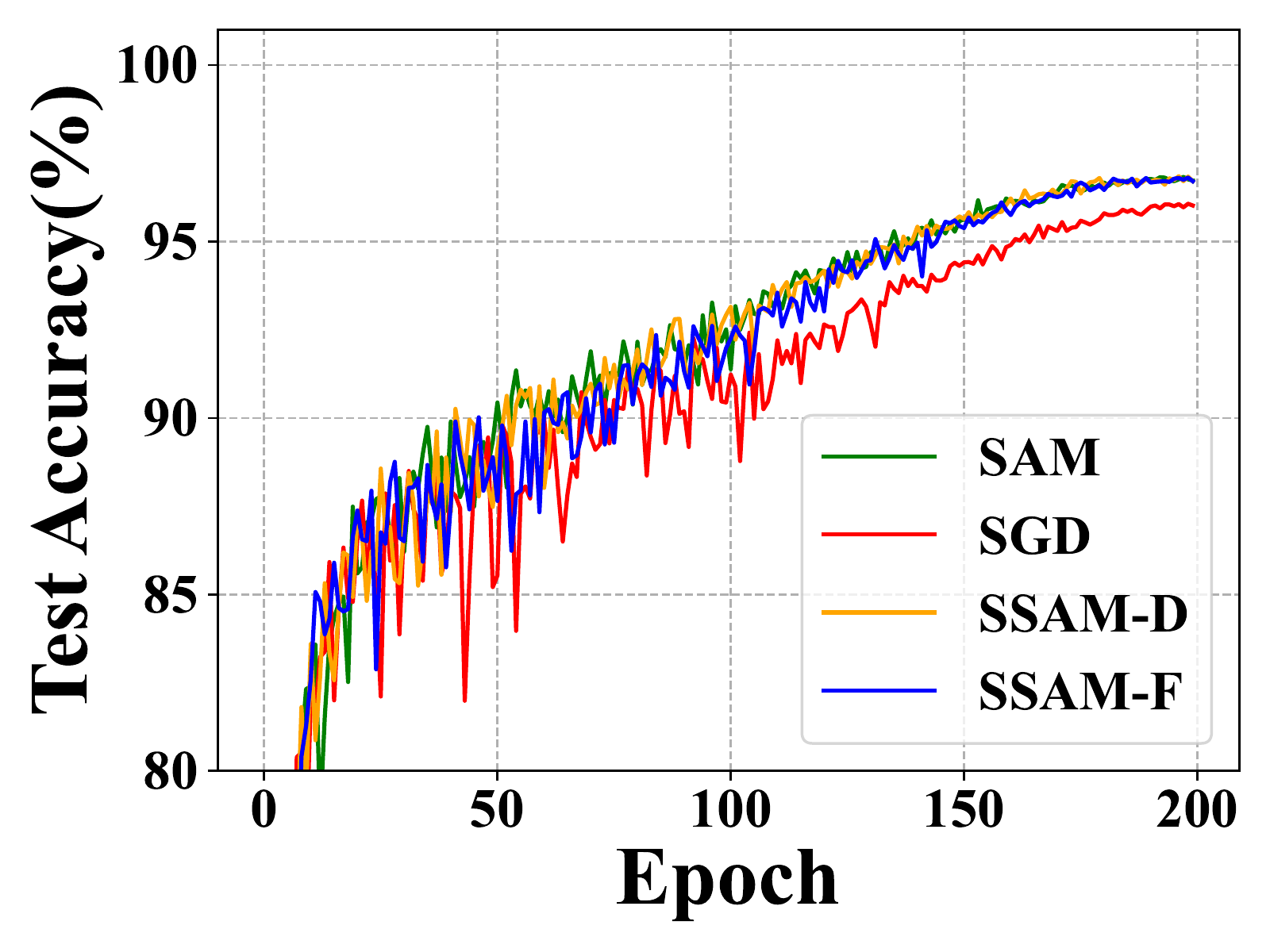}
    \label{fig:resnet18-cifar10-train-cruve}
}
\hspace{-5mm}
\subfigure[ResNet18 on CIFAR100]{
    \centering
    \includegraphics[width=0.32\linewidth]{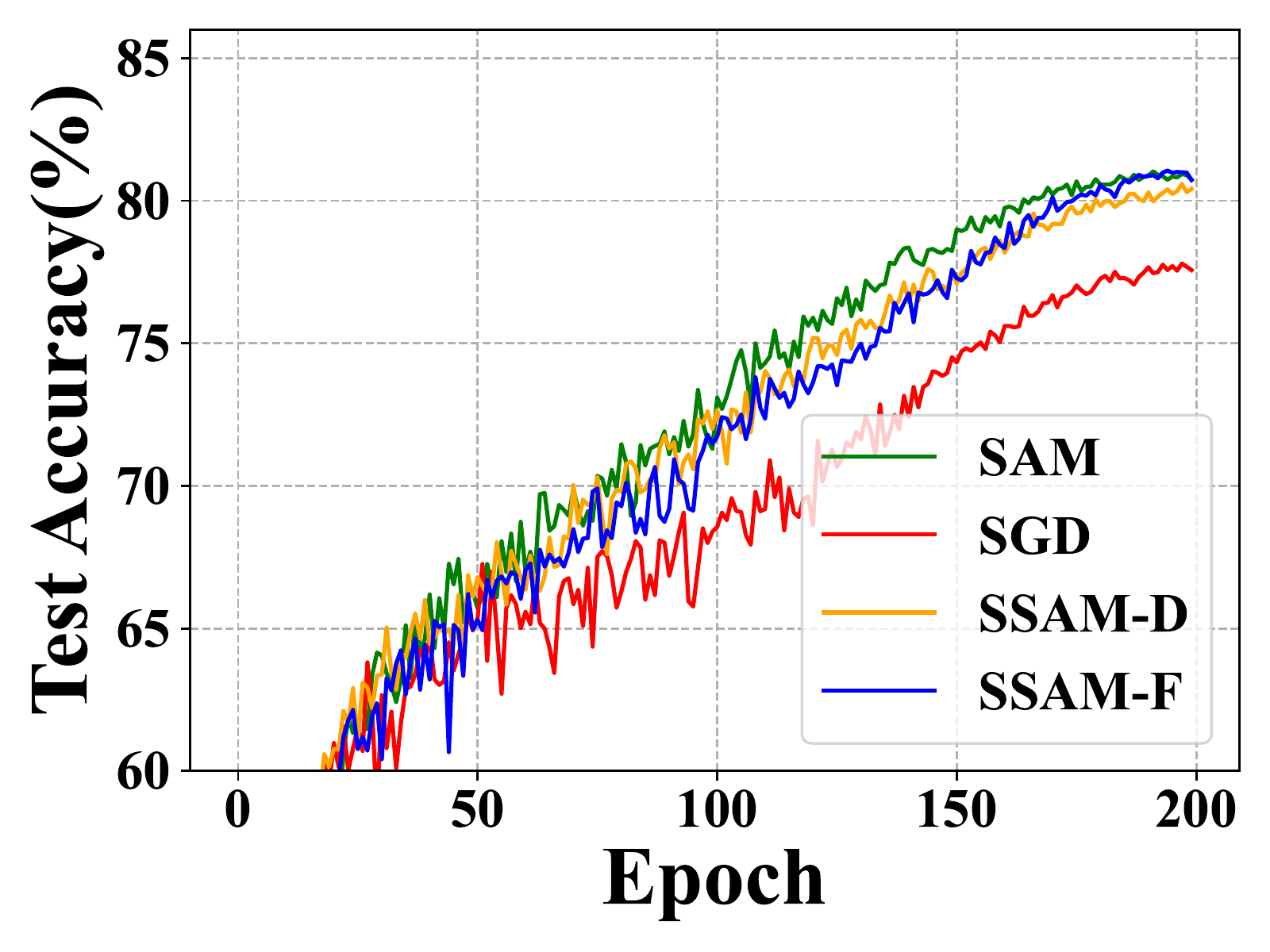}
    \label{fig:resnet18-cifar100-train-cruve}
}
\hspace{-5mm}
\subfigure[ResNet50 on ImageNet-1K]{
    \centering
    \includegraphics[width=0.32\linewidth]{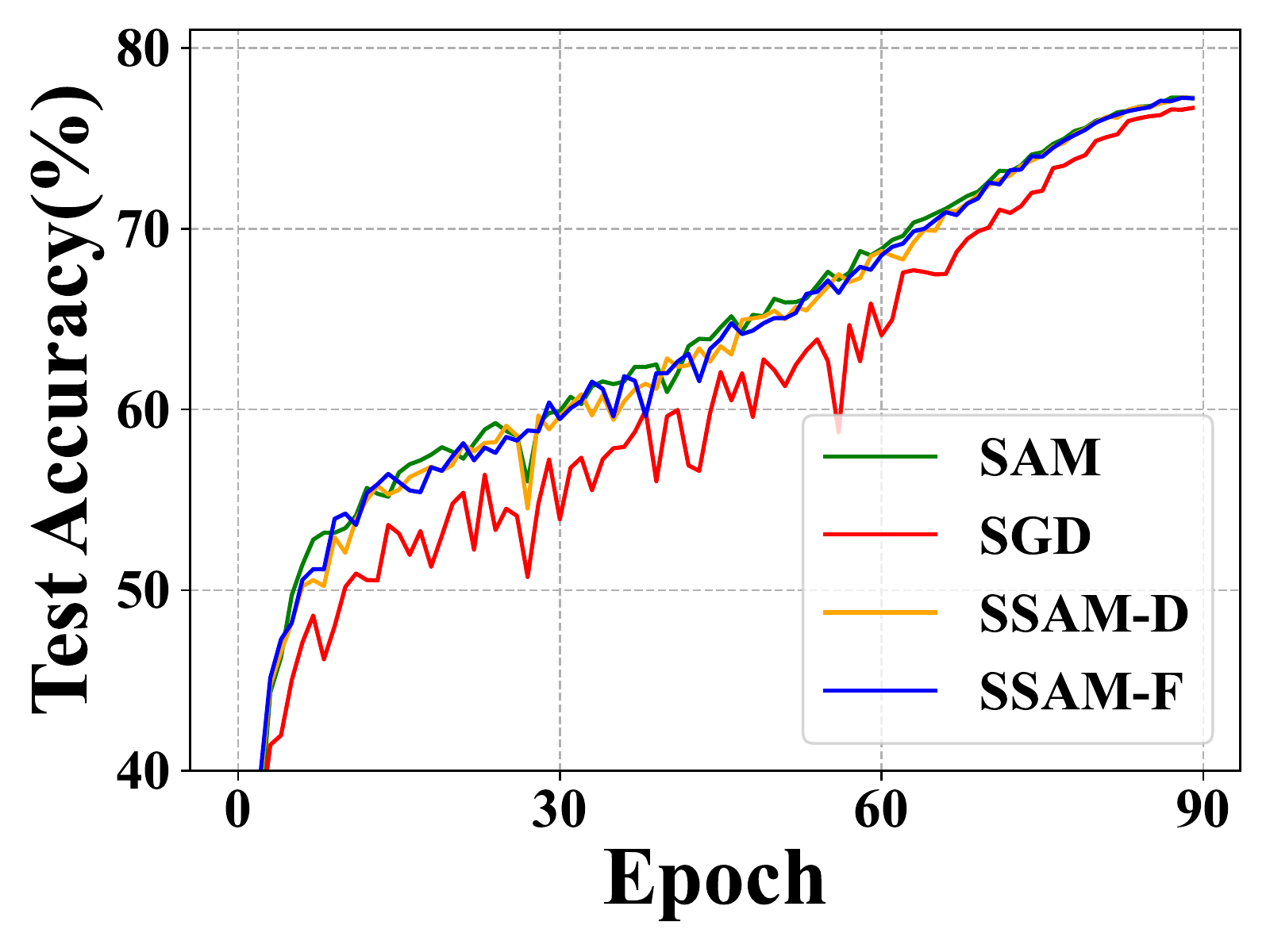}
    \label{fig:resnet50-imagenet-train-cruve}
}
\vspace{-3mm}
\caption{The training curves of SGD, SAM and SSAM. The sparsity of SSAM  is 50\%.}
\label[figure]{fig:train-curve}
\vspace{-0.3cm}
\end{figure}

\textbf{Training curves.}
We visualize the training curves of SGD, SAM and SSAM in~\cref{fig:train-curve}. The training curves of SGD are more jittery, while SAM is more stable. In SSAM, about half of gradient updates are the same as SGD, but its training curves are similar to those of SAM, suggesting the effectiveness.

\begin{figure}[ht]
\centering
\subfigure[WideResNet on CIFAR10.]{
    \centering
    \!\!\!
    \includegraphics[width=0.32\linewidth]{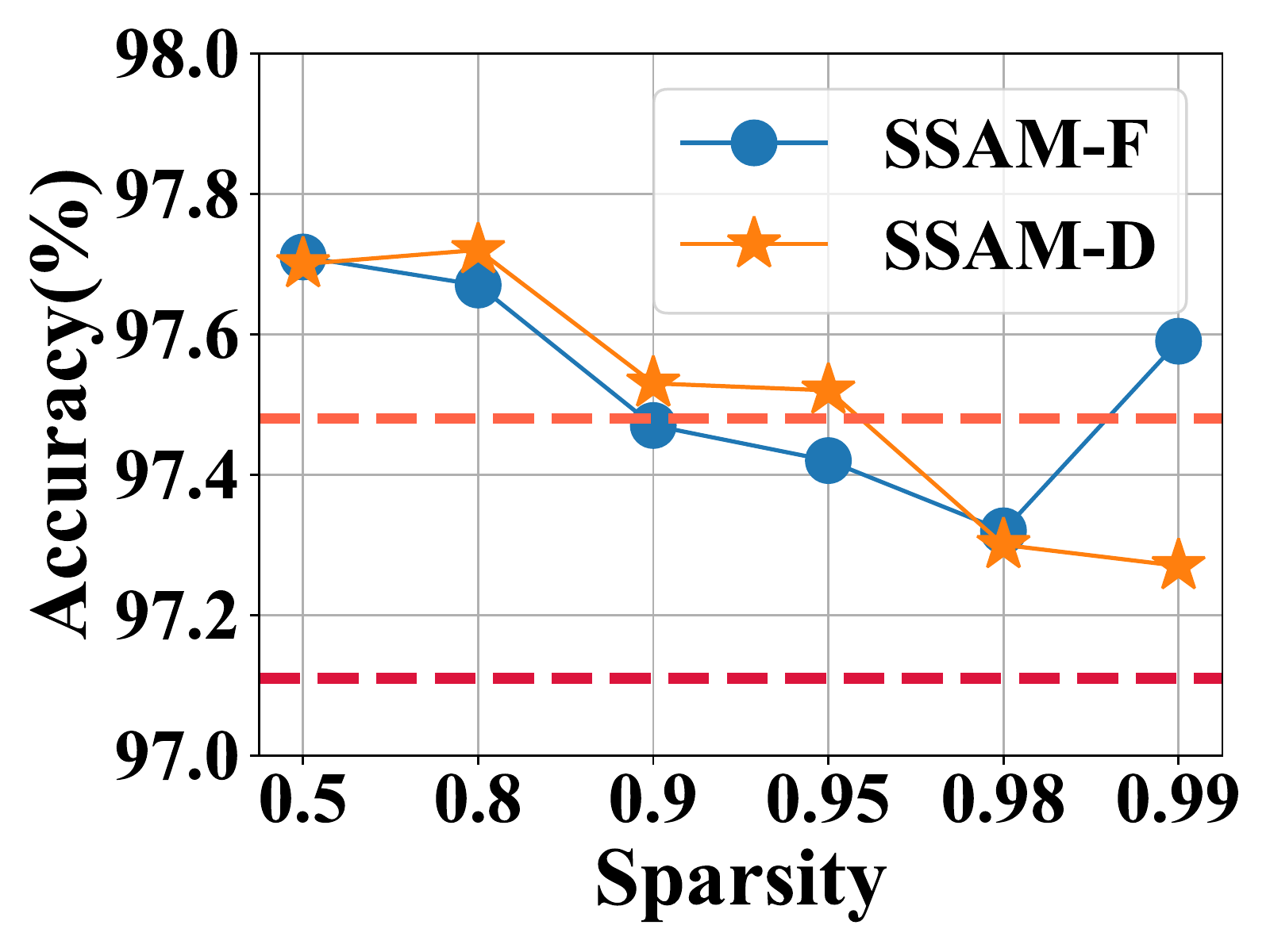}
}
\hspace{-3.3mm}
\subfigure[WideResNet on CIFAR100.]{
    \centering
    \includegraphics[width=0.32\linewidth]{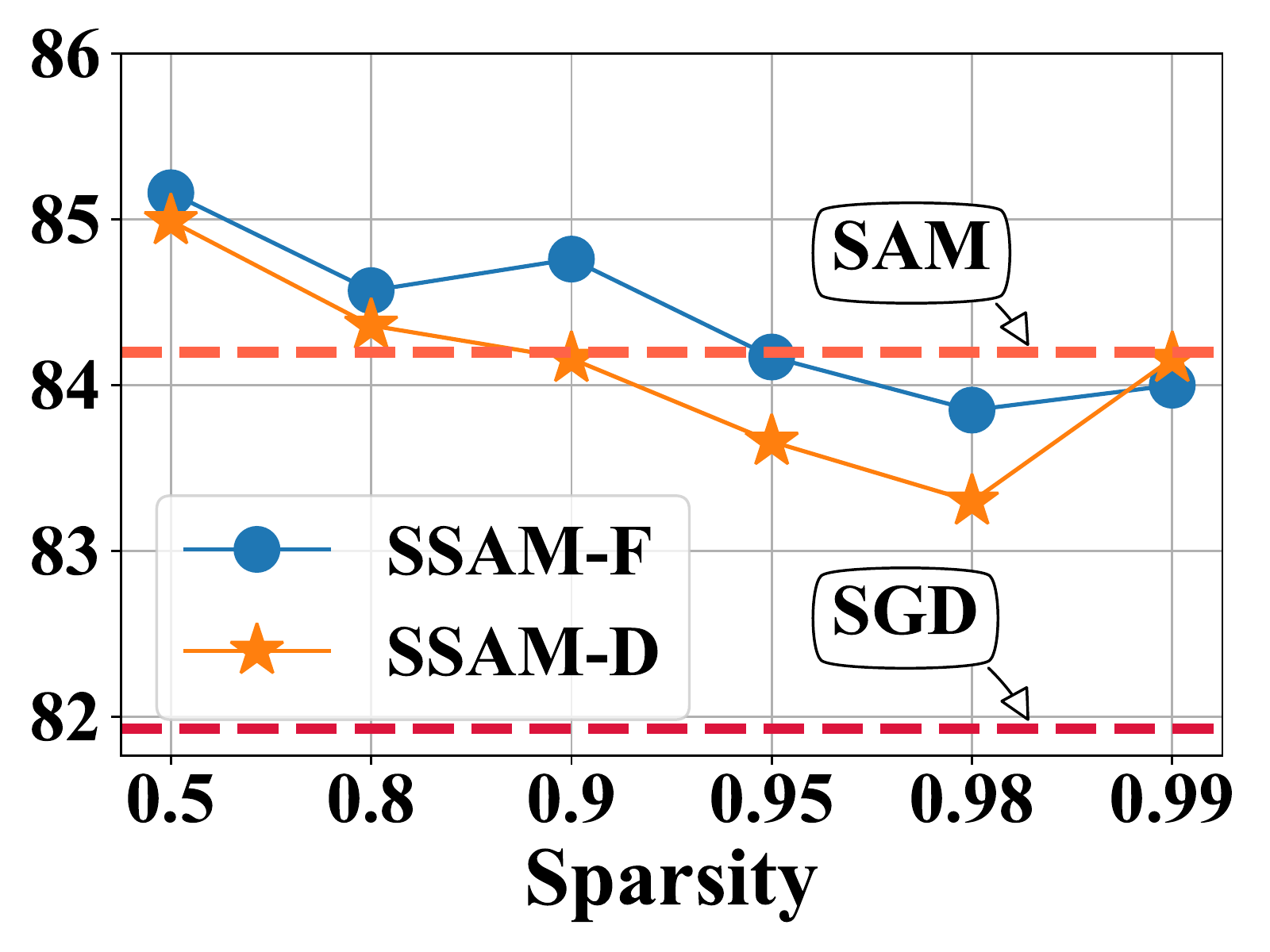}
}
\hspace{-3.3mm}
\subfigure[ResNet50 on ImageNet.]{
    \centering
    \includegraphics[width=0.32\linewidth]{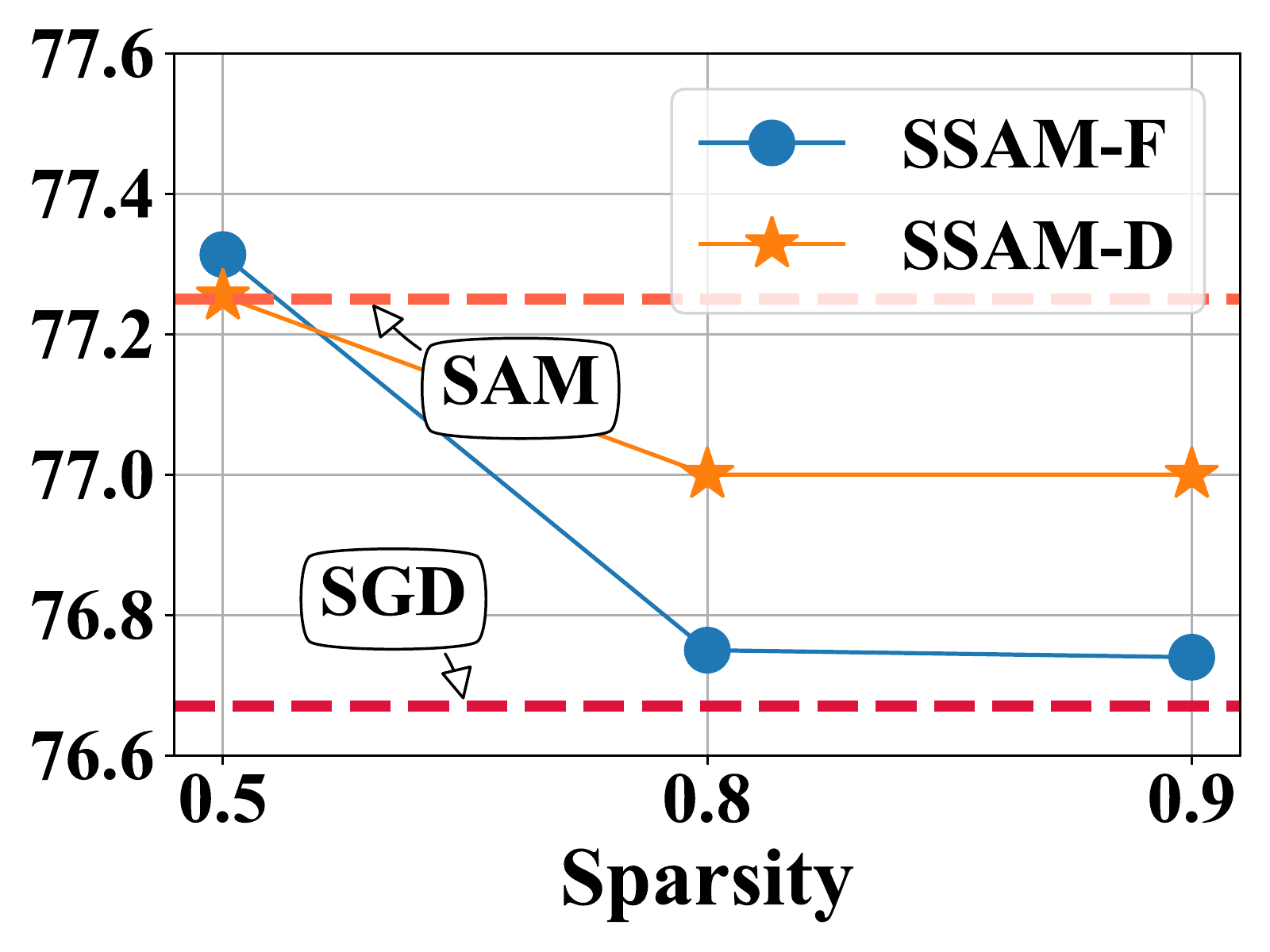}
    }
\vspace{-3.3mm}
\caption{Accuarcy \emph{v.s.} sparsity on CIFAR10, CIFAR100 and ImageNet datasets.}
\label[figure]{fig:sparsity-acc}
\end{figure}

\textbf{Sparsity \emph{vs.} Accuracy.}\ 
We report the effect of sparsity ratios in SSAM, as depicted in~\cref{fig:sparsity-acc}. We can observe that on CIFAR datasets, the sparsities of SSAM-F and SSAM-D pose little impact on performance. In addition, they can obtain better accuracies than SAM with up to 99\% sparsity. On the much larger dataset,~\emph{i.e.}, ImageNet, a higher sparsity will lead to more obvious performance drop. 

\subsection{SSAM with better generalization}
\begin{figure}[ht]
    \centering
    \includegraphics[width=1\linewidth]{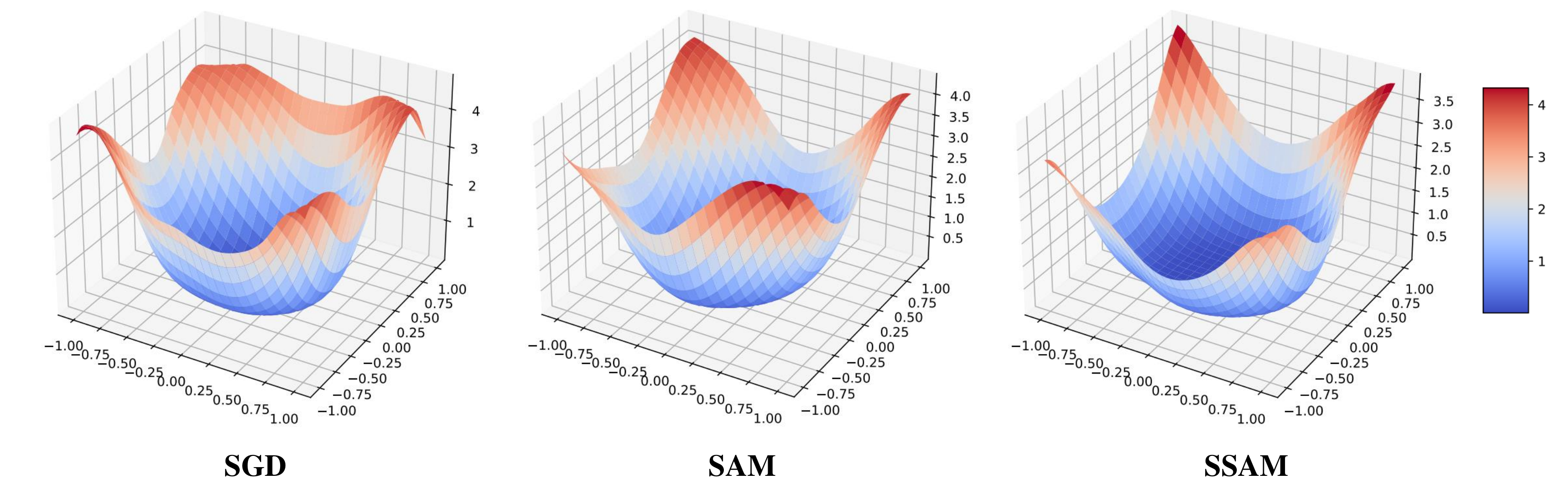}
    \caption{Training loss landscapes of ResNet18 on CIFAR10 trained with SGD, SAM, SSAM.}
    \label{fig:vis-landscape}
    \vspace{-0.4cm}
\end{figure}

\textbf{Visualization of landscape.} For a more intuitive comparison between different optimization schemes, we visualize the training loss landscapes of ResNet18 optimized by SGD, SAM and SSAM as shown in~\cref{fig:vis-landscape}. Following~\cite{vis-landscape}, we sample $50\times50$ points in the range of $[-1,1]$ from random "filter normalized"~\cite{vis-landscape} directions, \emph{i.e.}, the $x$ and $y$ axes. As shown in~\cref{fig:vis-landscape}, the landscape of SSAM is flatter than both SGD and SAM, and most of its area is low loss (blue). This result indicates that SSAM can smooth the loss landscape notably with sparse perturbation, and it also suggests that the complete perturbation on all parameters will result in suboptimal minima.

\begin{figure}[ht]
\centering
\subfigure[SGD Hessian spectrum]{
    \centering
    \includegraphics[width=0.32\linewidth]{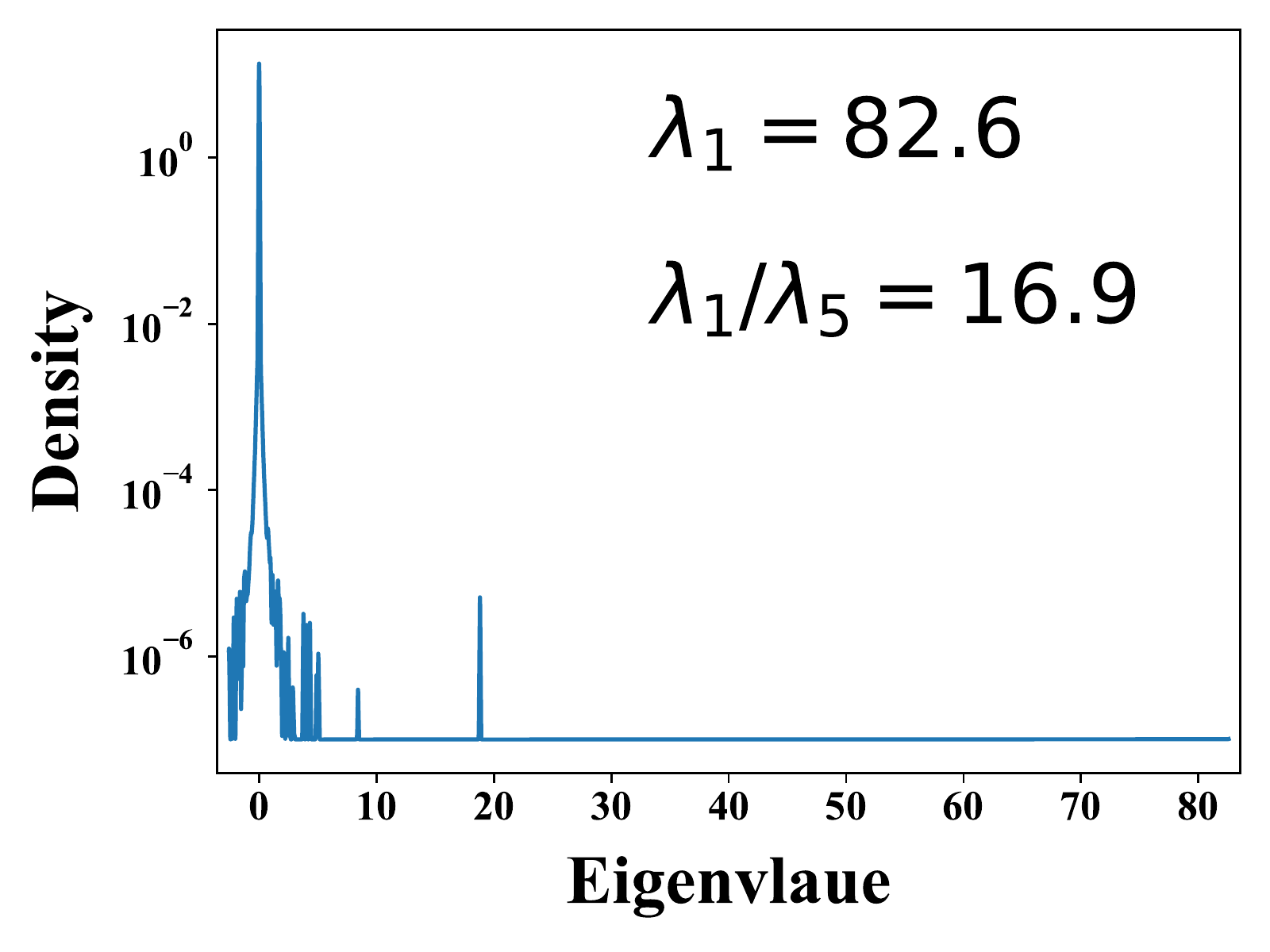}
    \label{fig:hessian-sgd}
}
\hspace{-5mm}
\subfigure[SAM Hessian spectrum]{
    \centering
    \includegraphics[width=0.32\linewidth]{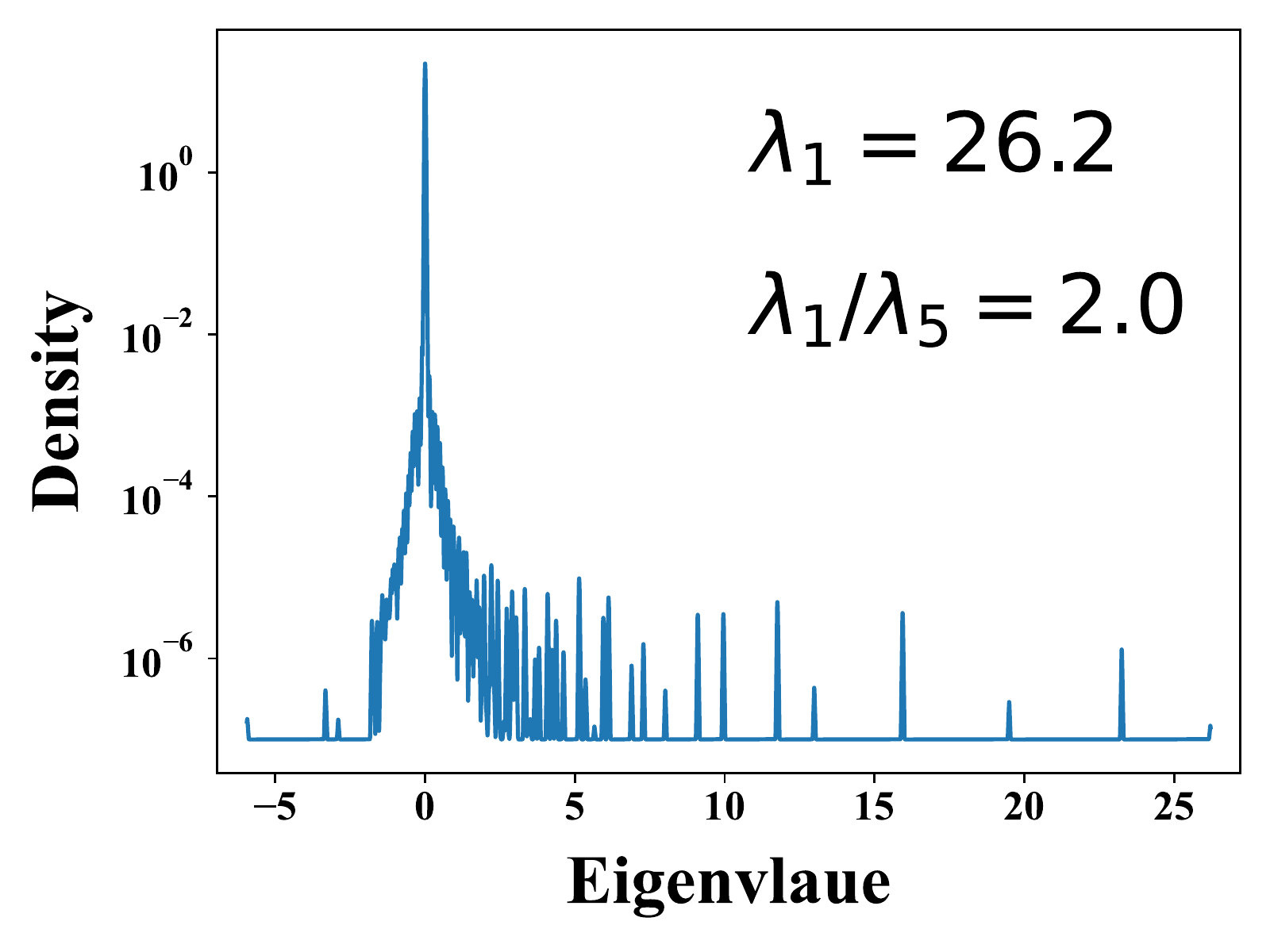}
    \label{fig:hessian-sam}
}
\hspace{-5mm}
\subfigure[SSAM Hessian spectrum]{
    \centering
    \includegraphics[width=0.32\linewidth]{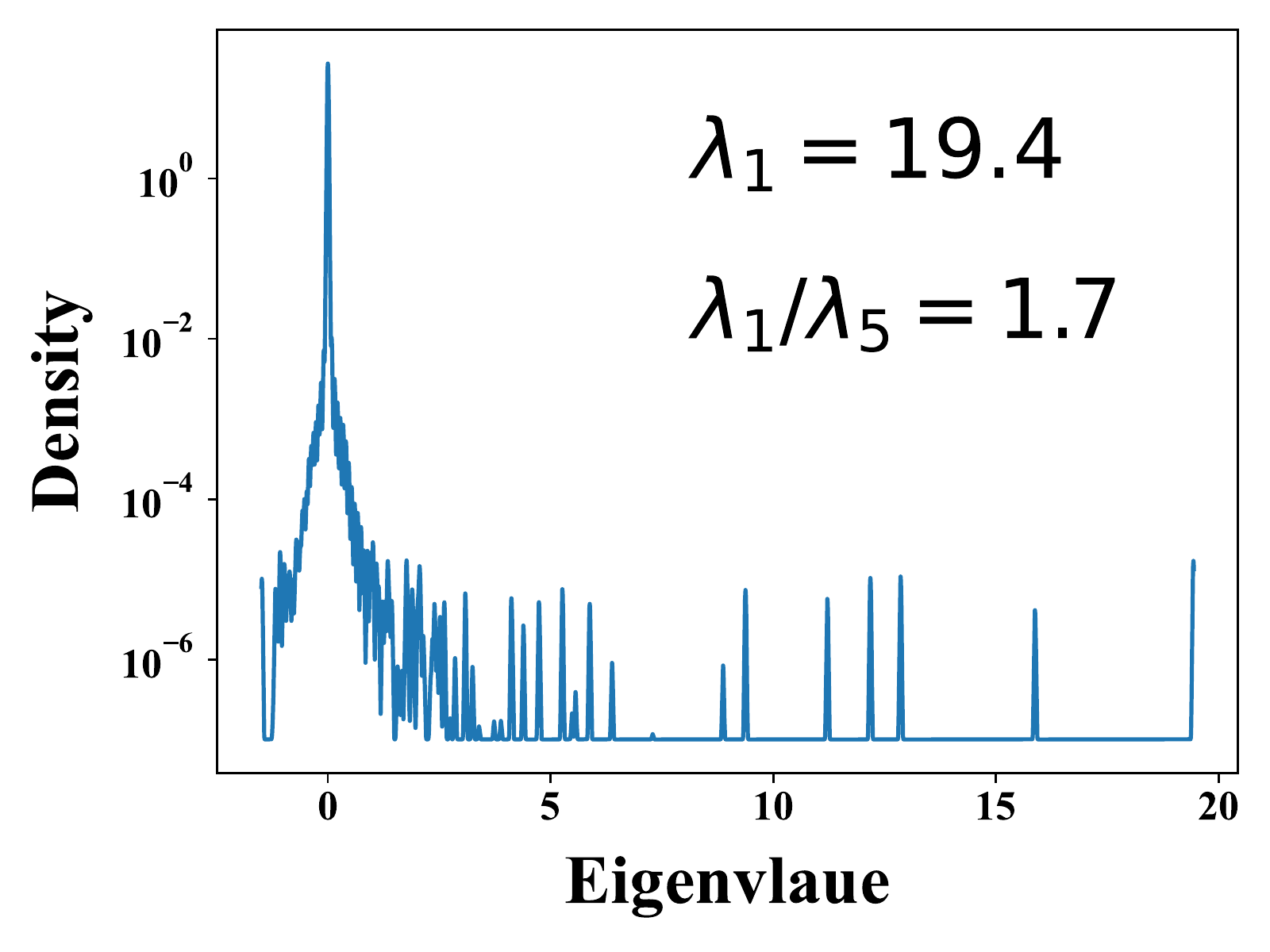}
    \label{fig:hessian-ssam}
}
\vspace{-3mm}
\caption{Hessian spectra of ResNet18 using SGD, SAM and SSAM on CIFAR10.}
\label{fig:hessian}
\vspace{-0.4cm}
\end{figure}

\textbf{Hessian spectra.}
In~\cref{fig:hessian}, we report the Hessian spectrum to demonstrate that SSAM can converge to a flat minima. Here, we also report the ratio of dominant eigenvalue to fifth largest ones,~\emph{i.e.}, 
$\lambda_1/\lambda_5$, used as the criteria in~\cite{sam, lambda1/lambda5}. 
We approximate the Hessian spectrum using the Lanczos algorithm~\cite{appro-hessian} and illustrate the Hessian spectra of ResNet18 using SGD, SAM and SSAM on CIFAR10. From this figure, we observe that the dominant eigenvalue $\lambda_1$ with SSAM is less than SGD and comparable to SAM. It confirms that SSAM with a sparse perturbation can still converge to the flat minima as SAM does, or even better. 

\section{Conclusion}
\vspace{-3mm}
In this paper, we reveal that the SAM gradients for most parameters are not significantly different from the SGD ones. Based on this finding, we propose an efficient training scheme called Sparse SAM, which is achieved by computing a sparse perturbation. We provide two solutions for sparse perturbation, which are based on Fisher information and dynamic sparse training, respectively.  In addition, we also theoretically prove that SSAM has the same convergence rate as SAM. We validate our SSAM on extensive datasets with various models. The experimental results show that retaining much better efficiency, SSAM can achieve competitive and even better performance than SAM.
\vspace{-3mm}
\section*{Acknowledgement}
\vspace{-3mm}
\small{
This work is supported by the Major Science and Technology Innovation 2030 “Brain Science and Brain-like Research” key project (No. 2021ZD0201405), the National Science Fund for Distinguished Young Scholars (No.
62025603), the National Natural Science Foundation of China (No. U21B2037, No. 62176222,
No. 62176223, No. 62176226, No. 62072386, No. 62072387, No. 62072389, and No. 62002305),
Guangdong Basic and Applied Basic Research Foundation (No. 2019B1515120049), and the Natural
Science Foundation of Fujian Province of China (No. 2021J01002).}
 

\newpage
{\small
\bibliographystyle{plainnat}
\bibliography{ref.bib}

\begin{thebibliography}{60}
\providecommand{\natexlab}[1]{#1}
\providecommand{\url}[1]{\texttt{#1}}
\expandafter\ifx\csname urlstyle\endcsname\relax
  \providecommand{\doi}[1]{doi: #1}\else
  \providecommand{\doi}{doi: \begingroup \urlstyle{rm}\Url}\fi

\bibitem[Andriushchenko and Flammarion(2021)]{understand-sam}
Maksym Andriushchenko and Nicolas Flammarion.
\newblock Understanding sharpness-aware minimization.
\newblock 2021.

\bibitem[Bellec et~al.(2017)Bellec, Kappel, Maass, and
  Legenstein]{dst-recent-deep-rewiring}
Guillaume Bellec, David Kappel, Wolfgang Maass, and Robert Legenstein.
\newblock Deep rewiring: Training very sparse deep networks.
\newblock \emph{arXiv preprint arXiv:1711.05136}, 2017.

\bibitem[Bottou(2010)]{bottou2010large}
L{\'e}on Bottou.
\newblock Large-scale machine learning with stochastic gradient descent.
\newblock In \emph{Proceedings of COMPSTAT'2010}, pages 177--186. Springer,
  2010.

\bibitem[Bottou et~al.(2018)Bottou, Curtis, and
  Nocedal]{bottou2018optimization}
L{\'e}on Bottou, Frank~E Curtis, and Jorge Nocedal.
\newblock Optimization methods for large-scale machine learning.
\newblock \emph{Siam Review}, 60\penalty0 (2):\penalty0 223--311, 2018.

\bibitem[Brock et~al.(2018)Brock, Donahue, and Simonyan]{biggan}
Andrew Brock, Jeff Donahue, and Karen Simonyan.
\newblock Large scale gan training for high fidelity natural image synthesis.
\newblock \emph{arXiv preprint arXiv:1809.11096}, 2018.

\bibitem[Chen et~al.(2022)Chen, Shen, Zou, and Liu]{chen2022towards}
Congliang Chen, Li~Shen, Fangyu Zou, and Wei Liu.
\newblock Towards practical adam: Non-convexity, convergence theory, and
  mini-batch acceleration.
\newblock \emph{Journal of Machine Learning Research}, 23:\penalty0 1--47,
  2022.

\bibitem[Chen et~al.(2021)Chen, Hsieh, and Gong]{vit-sam}
Xiangning Chen, Cho-Jui Hsieh, and Boqing Gong.
\newblock When vision transformers outperform resnets without pre-training or
  strong data augmentations.
\newblock \emph{arXiv preprint arXiv:2106.01548}, 2021.

\bibitem[Deng et~al.(2009)Deng, Dong, Socher, Li, Li, and Fei-Fei]{imagenet}
Jia Deng, Wei Dong, Richard Socher, Li-Jia Li, Kai Li, and Li~Fei-Fei.
\newblock Imagenet: A large-scale hierarchical image database.
\newblock In \emph{2009 IEEE conference on computer vision and pattern
  recognition}, pages 248--255. Ieee, 2009.

\bibitem[Dettmers and Zettlemoyer(2019)]{sparse-cosine}
Tim Dettmers and Luke Zettlemoyer.
\newblock Sparse networks from scratch: Faster training without losing
  performance.
\newblock \emph{arXiv preprint arXiv:1907.04840}, 2019.

\bibitem[Devlin et~al.(2018)Devlin, Chang, Lee, and Toutanova]{bert}
Jacob Devlin, Ming-Wei Chang, Kenton Lee, and Kristina Toutanova.
\newblock Bert: Pre-training of deep bidirectional transformers for language
  understanding.
\newblock \emph{arXiv preprint arXiv:1810.04805}, 2018.

\bibitem[DeVries and Taylor(2017)]{cutout}
Terrance DeVries and Graham~W Taylor.
\newblock Improved regularization of convolutional neural networks with cutout.
\newblock \emph{arXiv preprint arXiv:1708.04552}, 2017.

\bibitem[Dinh et~al.(2017)Dinh, Pascanu, Bengio, and
  Bengio]{sharp-can-generalize}
Laurent Dinh, Razvan Pascanu, Samy Bengio, and Yoshua Bengio.
\newblock Sharp minima can generalize for deep nets.
\newblock In \emph{International Conference on Machine Learning}, pages
  1019--1028. PMLR, 2017.

\bibitem[Dosovitskiy et~al.(2020)Dosovitskiy, Beyer, Kolesnikov, Weissenborn,
  Zhai, Unterthiner, Dehghani, Minderer, Heigold, Gelly, et~al.]{vit}
Alexey Dosovitskiy, Lucas Beyer, Alexander Kolesnikov, Dirk Weissenborn,
  Xiaohua Zhai, Thomas Unterthiner, Mostafa Dehghani, Matthias Minderer, Georg
  Heigold, Sylvain Gelly, et~al.
\newblock An image is worth 16x16 words: Transformers for image recognition at
  scale.
\newblock \emph{arXiv preprint arXiv:2010.11929}, 2020.

\bibitem[Du et~al.(2021)Du, Yan, Feng, Zhou, Zhen, Goh, and Tan]{esam}
Jiawei Du, Hanshu Yan, Jiashi Feng, Joey~Tianyi Zhou, Liangli Zhen, Rick
  Siow~Mong Goh, and Vincent~YF Tan.
\newblock Efficient sharpness-aware minimization for improved training of
  neural networks.
\newblock \emph{arXiv preprint arXiv:2110.03141}, 2021.

\bibitem[Evci et~al.(2020)Evci, Gale, Menick, Castro, and Elsen]{rig}
Utku Evci, Trevor Gale, Jacob Menick, Pablo~Samuel Castro, and Erich Elsen.
\newblock Rigging the lottery: Making all tickets winners.
\newblock In \emph{International Conference on Machine Learning}, pages
  2943--2952. PMLR, 2020.

\bibitem[Fisher(1922)]{fisher-information}
Ronald~A Fisher.
\newblock On the mathematical foundations of theoretical statistics.
\newblock \emph{Philosophical transactions of the Royal Society of London.
  Series A, containing papers of a mathematical or physical character},
  222\penalty0 (594-604):\penalty0 309--368, 1922.

\bibitem[Foret et~al.(2020)Foret, Kleiner, Mobahi, and Neyshabur]{sam}
Pierre Foret, Ariel Kleiner, Hossein Mobahi, and Behnam Neyshabur.
\newblock Sharpness-aware minimization for efficiently improving
  generalization.
\newblock In \emph{International Conference on Learning Representations}, 2020.

\bibitem[Frankle and Carbin(2018)]{LTH}
Jonathan Frankle and Michael Carbin.
\newblock The lottery ticket hypothesis: Finding sparse, trainable neural
  networks.
\newblock \emph{arXiv preprint arXiv:1803.03635}, 2018.

\bibitem[George et~al.(2018)George, Laurent, Bouthillier, Ballas, and
  Vincent]{KFAC}
Thomas George, C{\'e}sar Laurent, Xavier Bouthillier, Nicolas Ballas, and
  Pascal Vincent.
\newblock Fast approximate natural gradient descent in a kronecker factored
  eigenbasis.
\newblock \emph{Advances in Neural Information Processing Systems}, 31, 2018.

\bibitem[Ghadimi and Lan(2013)]{ghadimi2013stochastic}
Saeed Ghadimi and Guanghui Lan.
\newblock Stochastic first-and zeroth-order methods for nonconvex stochastic
  programming.
\newblock \emph{SIAM Journal on Optimization}, 23\penalty0 (4):\penalty0
  2341--2368, 2013.

\bibitem[Ghorbani et~al.(2019)Ghorbani, Krishnan, and Xiao]{appro-hessian}
Behrooz Ghorbani, Shankar Krishnan, and Ying Xiao.
\newblock An investigation into neural net optimization via hessian eigenvalue
  density.
\newblock In \emph{International Conference on Machine Learning}, pages
  2232--2241. PMLR, 2019.

\bibitem[Grosse and Martens(2016)]{KFAC-cnn}
Roger Grosse and James Martens.
\newblock A kronecker-factored approximate fisher matrix for convolution
  layers.
\newblock In \emph{International Conference on Machine Learning}, pages
  573--582. PMLR, 2016.

\bibitem[Han et~al.(2015)Han, Mao, and Dally]{prune-pipeline}
Song Han, Huizi Mao, and William~J Dally.
\newblock Deep compression: Compressing deep neural networks with pruning,
  trained quantization and huffman coding.
\newblock \emph{arXiv preprint arXiv:1510.00149}, 2015.

\bibitem[Hansen et~al.(1997)Hansen, Larsen, and Fog]{early-stop}
Lars~Kai Hansen, Jan Larsen, and Torben Fog.
\newblock Early stop criterion from the bootstrap ensemble.
\newblock In \emph{1997 IEEE International Conference on Acoustics, Speech, and
  Signal Processing}, volume~4, pages 3205--3208. IEEE, 1997.

\bibitem[He et~al.(2016)He, Zhang, Ren, and Sun]{resnet}
Kaiming He, Xiangyu Zhang, Shaoqing Ren, and Jian Sun.
\newblock Deep residual learning for image recognition.
\newblock In \emph{Proceedings of the IEEE conference on computer vision and
  pattern recognition}, pages 770--778, 2016.

\bibitem[Hinton et~al.(2012)Hinton, Srivastava, Krizhevsky, Sutskever, and
  Salakhutdinov]{dropout}
Geoffrey~E Hinton, Nitish Srivastava, Alex Krizhevsky, Ilya Sutskever, and
  Ruslan~R Salakhutdinov.
\newblock Improving neural networks by preventing co-adaptation of feature
  detectors.
\newblock \emph{arXiv preprint arXiv:1207.0580}, 2012.

\bibitem[Hochreiter and Schmidhuber(1994)]{flat-minima}
Sepp Hochreiter and J{\"u}rgen Schmidhuber.
\newblock Simplifying neural nets by discovering flat minima.
\newblock \emph{Advances in neural information processing systems}, 7, 1994.

\bibitem[Ioffe and Szegedy(2015)]{bn}
Sergey Ioffe and Christian Szegedy.
\newblock Batch normalization: Accelerating deep network training by reducing
  internal covariate shift.
\newblock In \emph{International conference on machine learning}, pages
  448--456. PMLR, 2015.

\bibitem[Jastrzebski et~al.(2020)Jastrzebski, Szymczak, Fort, Arpit, Tabor,
  Cho, and Geras]{lambda1/lambda5}
Stanislaw Jastrzebski, Maciej Szymczak, Stanislav Fort, Devansh Arpit, Jacek
  Tabor, Kyunghyun Cho, and Krzysztof Geras.
\newblock The break-even point on optimization trajectories of deep neural
  networks.
\newblock \emph{arXiv preprint arXiv:2002.09572}, 2020.

\bibitem[Jayakumar et~al.(2020)Jayakumar, Pascanu, Rae, Osindero, and
  Elsen]{dst-recent-topkast}
Siddhant Jayakumar, Razvan Pascanu, Jack Rae, Simon Osindero, and Erich Elsen.
\newblock Top-kast: Top-k always sparse training.
\newblock \emph{Advances in Neural Information Processing Systems},
  33:\penalty0 20744--20754, 2020.

\bibitem[Keskar et~al.(2016)Keskar, Mudigere, Nocedal, Smelyanskiy, and
  Tang]{LBtrain}
Nitish~Shirish Keskar, Dheevatsa Mudigere, Jorge Nocedal, Mikhail Smelyanskiy,
  and Ping Tak~Peter Tang.
\newblock On large-batch training for deep learning: Generalization gap and
  sharp minima.
\newblock 2016.

\bibitem[Kirkpatrick et~al.(2017)Kirkpatrick, Pascanu, Rabinowitz, Veness,
  Desjardins, Rusu, Milan, Quan, Ramalho, Grabska-Barwinska,
  et~al.]{pnas-forgetting-in-neural-networks}
James Kirkpatrick, Razvan Pascanu, Neil Rabinowitz, Joel Veness, Guillaume
  Desjardins, Andrei~A Rusu, Kieran Milan, John Quan, Tiago Ramalho, Agnieszka
  Grabska-Barwinska, et~al.
\newblock Overcoming catastrophic forgetting in neural networks.
\newblock \emph{Proceedings of the national academy of sciences}, 114\penalty0
  (13):\penalty0 3521--3526, 2017.

\bibitem[Krizhevsky et~al.(2009)]{cifar10/100}
Alex Krizhevsky et~al.
\newblock Learning multiple layers of features from tiny images.
\newblock 2009.

\bibitem[Kwon et~al.(2021)Kwon, Kim, Park, and Choi]{asam}
Jungmin Kwon, Jeongseop Kim, Hyunseo Park, and In~Kwon Choi.
\newblock Asam: Adaptive sharpness-aware minimization for scale-invariant
  learning of deep neural networks.
\newblock In \emph{International Conference on Machine Learning}, pages
  5905--5914. PMLR, 2021.

\bibitem[LeCun et~al.(1989)LeCun, Denker, and Solla]{lecun-1th-prune}
Yann LeCun, John Denker, and Sara Solla.
\newblock Optimal brain damage.
\newblock \emph{Advances in neural information processing systems}, 2, 1989.

\bibitem[Li et~al.(2018)Li, Xu, Taylor, Studer, and Goldstein]{vis-landscape}
Hao Li, Zheng Xu, Gavin Taylor, Christoph Studer, and Tom Goldstein.
\newblock Visualizing the loss landscape of neural nets.
\newblock \emph{Advances in neural information processing systems}, 31, 2018.

\bibitem[Liu et~al.(2021{\natexlab{a}})Liu, Chen, Chen, Atashgahi, Yin, Kou,
  Shen, Pechenizkiy, Wang, and Mocanu]{liu2021sparse}
Shiwei Liu, Tianlong Chen, Xiaohan Chen, Zahra Atashgahi, Lu~Yin, Huanyu Kou,
  Li~Shen, Mykola Pechenizkiy, Zhangyang Wang, and Decebal~Constantin Mocanu.
\newblock Sparse training via boosting pruning plasticity with
  neuroregeneration.
\newblock \emph{Advances in Neural Information Processing Systems},
  34:\penalty0 9908--9922, 2021{\natexlab{a}}.

\bibitem[Liu et~al.(2021{\natexlab{b}})Liu, Chen, Chen, Shen, Mocanu, Wang, and
  Pechenizkiy]{liu2022unreasonable}
Shiwei Liu, Tianlong Chen, Xiaohan Chen, Li~Shen, Decebal~Constantin Mocanu,
  Zhangyang Wang, and Mykola Pechenizkiy.
\newblock The unreasonable effectiveness of random pruning: Return of the most
  naive baseline for sparse training.
\newblock In \emph{International Conference on Learning Representations},
  2021{\natexlab{b}}.

\bibitem[Liu et~al.(2021{\natexlab{c}})Liu, Chen, Cheng, Hsieh, and
  You]{con-adv}
Yong Liu, Xiangning Chen, Minhao Cheng, Cho-Jui Hsieh, and Yang You.
\newblock Concurrent adversarial learning for large-batch training.
\newblock \emph{arXiv preprint arXiv:2106.00221}, 2021{\natexlab{c}}.

\bibitem[Liu et~al.(2021{\natexlab{d}})Liu, Lin, Cao, Hu, Wei, Zhang, Lin, and
  Guo]{swin-transformer}
Ze~Liu, Yutong Lin, Yue Cao, Han Hu, Yixuan Wei, Zheng Zhang, Stephen Lin, and
  Baining Guo.
\newblock Swin transformer: Hierarchical vision transformer using shifted
  windows.
\newblock In \emph{Proceedings of the IEEE/CVF International Conference on
  Computer Vision}, pages 10012--10022, 2021{\natexlab{d}}.

\bibitem[Martens and Grosse(2015)]{KFAC-opt}
James Martens and Roger Grosse.
\newblock Optimizing neural networks with kronecker-factored approximate
  curvature.
\newblock In \emph{International conference on machine learning}, pages
  2408--2417. PMLR, 2015.

\bibitem[Mnih et~al.(2016)Mnih, Badia, Mirza, Graves, Lillicrap, Harley,
  Silver, and Kavukcuoglu]{entropy-regularization}
Volodymyr Mnih, Adria~Puigdomenech Badia, Mehdi Mirza, Alex Graves, Timothy
  Lillicrap, Tim Harley, David Silver, and Koray Kavukcuoglu.
\newblock Asynchronous methods for deep reinforcement learning.
\newblock In \emph{International conference on machine learning}, pages
  1928--1937. PMLR, 2016.

\bibitem[Mostafa and Wang(2019)]{dst-recent-reparameterization}
Hesham Mostafa and Xin Wang.
\newblock Parameter efficient training of deep convolutional neural networks by
  dynamic sparse reparameterization.
\newblock In \emph{International Conference on Machine Learning}, pages
  4646--4655. PMLR, 2019.

\bibitem[Neyshabur et~al.(2017)Neyshabur, Bhojanapalli, McAllester, and
  Srebro]{exploring-generalization}
Behnam Neyshabur, Srinadh Bhojanapalli, David McAllester, and Nati Srebro.
\newblock Exploring generalization in deep learning.
\newblock \emph{Advances in neural information processing systems}, 30, 2017.

\bibitem[nvidia ampere architecture()]{nvidia-ampere-tensor-core}
nvidia ampere architecture.
\newblock Nvidia ampere tensor core.
\newblock \url{https://www.nvidia.com/en-us/data-center/ampere-architecture/}.

\bibitem[nvidia auto sparsity()]{nvidia-asp}
nvidia auto sparsity.
\newblock Accelerating sparsity in the nvidia ampere architecture.
\newblock
  \url{https://developer.download.nvidia.com/video/gputechconf/gtc/2020/presentations/s22085-accelerating-sparsity-in-the-nvidia-ampere-architecture\%E2\%80\%8B.pdf}.

\bibitem[nvidia whitepaper()]{nvidia-whitepaper}
nvidia whitepaper.
\newblock Nvidia a100 tensor core gpu architecture.
\newblock
  \url{https://images.nvidia.com/aem-dam/en-zz/Solutions/data-center/nvidia-ampere-architecture-whitepaper.pdf}.

\bibitem[Pytorch Cifar100()]{pytorch-cifar100-github}
Pytorch Cifar100.
\newblock Pytorch cifar100 github.
\newblock \url{https://github.com/weiaicunzai/pytorch-cifar100}.

\bibitem[Singh and Alistarh(2020)]{woodfisher}
Sidak~Pal Singh and Dan Alistarh.
\newblock Woodfisher: Efficient second-order approximation for neural network
  compression.
\newblock \emph{Advances in Neural Information Processing Systems},
  33:\penalty0 18098--18109, 2020.

\bibitem[Sung et~al.(2021)Sung, Nair, and Raffel]{fisher-mask}
Yi-Lin Sung, Varun Nair, and Colin~A Raffel.
\newblock Training neural networks with fixed sparse masks.
\newblock \emph{Advances in Neural Information Processing Systems}, 34, 2021.

\bibitem[Szegedy et~al.(2016)Szegedy, Vanhoucke, Ioffe, Shlens, and
  Wojna]{label-smoothing}
Christian Szegedy, Vincent Vanhoucke, Sergey Ioffe, Jon Shlens, and Zbigniew
  Wojna.
\newblock Rethinking the inception architecture for computer vision.
\newblock In \emph{Proceedings of the IEEE conference on computer vision and
  pattern recognition}, pages 2818--2826, 2016.

\bibitem[Theis et~al.(2018)Theis, Korshunova, Tejani, and
  Husz{\'a}r]{fisher-pruning}
Lucas Theis, Iryna Korshunova, Alykhan Tejani, and Ferenc Husz{\'a}r.
\newblock Faster gaze prediction with dense networks and fisher pruning.
\newblock \emph{arXiv preprint arXiv:1801.05787}, 2018.

\bibitem[Vaswani et~al.(2017)Vaswani, Shazeer, Parmar, Uszkoreit, Jones, Gomez,
  Kaiser, and Polosukhin]{transformer}
Ashish Vaswani, Noam Shazeer, Niki Parmar, Jakob Uszkoreit, Llion Jones,
  Aidan~N Gomez, {\L}ukasz Kaiser, and Illia Polosukhin.
\newblock Attention is all you need.
\newblock \emph{Advances in neural information processing systems}, 30, 2017.

\bibitem[Wu et~al.(2020)Wu, Xia, and Wang]{adversarial-robust}
Dongxian Wu, Shu-Tao Xia, and Yisen Wang.
\newblock Adversarial weight perturbation helps robust generalization.
\newblock \emph{Advances in Neural Information Processing Systems},
  33:\penalty0 2958--2969, 2020.

\bibitem[Xu et~al.(2022)Xu, He, Cheng, Wang, and Cheng]{2:4-truly-sparse}
Weixiang Xu, Xiangyu He, Ke~Cheng, Peisong Wang, and Jian Cheng.
\newblock Towards fully sparse training: Information restoration with spatial
  similarity.
\newblock 2022.

\bibitem[Yun et~al.(2019)Yun, Han, Oh, Chun, Choe, and Yoo]{cutmix}
Sangdoo Yun, Dongyoon Han, Seong~Joon Oh, Sanghyuk Chun, Junsuk Choe, and
  Youngjoon Yoo.
\newblock Cutmix: Regularization strategy to train strong classifiers with
  localizable features.
\newblock In \emph{Proceedings of the IEEE/CVF international conference on
  computer vision}, pages 6023--6032, 2019.

\bibitem[Zagoruyko and Komodakis(2016)]{wideresnet}
Sergey Zagoruyko and Nikos Komodakis.
\newblock Wide residual networks.
\newblock \emph{arXiv preprint arXiv:1605.07146}, 2016.

\bibitem[Zhang et~al.(2017)Zhang, Cisse, Dauphin, and Lopez-Paz]{mixup}
Hongyi Zhang, Moustapha Cisse, Yann~N Dauphin, and David Lopez-Paz.
\newblock mixup: Beyond empirical risk minimization.
\newblock \emph{arXiv preprint arXiv:1710.09412}, 2017.

\bibitem[Zhao et~al.(2022)Zhao, Zhang, and Hu]{penalize-sam}
Yang Zhao, Hao Zhang, and Xiuyuan Hu.
\newblock Penalizing gradient norm for efficiently improving generalization in
  deep learning.
\newblock \emph{arXiv preprint arXiv:2202.03599}, 2022.

\bibitem[Zhuang et~al.(2022)Zhuang, Gong, Yuan, Cui, Adam, Dvornek, Tatikonda,
  Duncan, and Liu]{surrogate}
Juntang Zhuang, Boqing Gong, Liangzhe Yuan, Yin Cui, Hartwig Adam, Nicha
  Dvornek, Sekhar Tatikonda, James Duncan, and Ting Liu.
\newblock Surrogate gap minimization improves sharpness-aware training.
\newblock \emph{arXiv preprint arXiv:2203.08065}, 2022.

\end{thebibliography}
}

\newpage
\section*{Checklist}

\begin{enumerate}

\item For all authors...
\begin{enumerate}
  \item Do the main claims made in the abstract and introduction accurately reflect the paper's contributions and scope?
    \answerYes{}
  \item Did you describe the limitations of your work?
    \answerNo{}
  \item Did you discuss any potential negative societal impacts of your work?
    \answerNo{}
  \item Have you read the ethics review guidelines and ensured that your paper conforms to them?
    \answerYes{}
\end{enumerate}

\item If you are including theoretical results...
\begin{enumerate}
  \item Did you state the full set of assumptions of all theoretical results?
    \answerYes{See Section~\ref{sec:method:theoretical}.}
        \item Did you include complete proofs of all theoretical results?
    \answerYes{The detailed proof is in Appendix.}
\end{enumerate}

\item If you ran experiments...
\begin{enumerate}
  \item Did you include the code, data, and instructions needed to reproduce the main experimental results (either in the supplemental material or as a URL)?
    \answerYes{}
  \item Did you specify all the training details (e.g., data splits, hyperparameters, how they were chosen)?
    \answerYes{See Section~\ref{sec:experiments-implement-detail}.}
        \item Did you report error bars (e.g., with respect to the random seed after running experiments multiple times)?
    \answerNo{}
        \item Did you include the total amount of compute and the type of resources used (e.g., type of GPUs, internal cluster, or cloud provider)?
    \answerNo{}
\end{enumerate}

\item If you are using existing assets (e.g., code, data, models) or curating/releasing new assets...
\begin{enumerate}
  \item If your work uses existing assets, did you cite the creators?
    \answerNA{}
  \item Did you mention the license of the assets?
    \answerNA{}
  \item Did you include any new assets either in the supplemental material or as a URL?
    \answerNA{}
  \item Did you discuss whether and how consent was obtained from people whose data you're using/curating?
    \answerNA{}
  \item Did you discuss whether the data you are using/curating contains personally identifiable information or offensive content?
    \answerNA{}
\end{enumerate}

\item If you used crowdsourcing or conducted research with human subjects...
\begin{enumerate}
  \item Did you include the full text of instructions given to participants and screenshots, if applicable?
    \answerNA{}
  \item Did you describe any potential participant risks, with links to Institutional Review Board (IRB) approvals, if applicable?
    \answerNA{}
  \item Did you include the estimated hourly wage paid to participants and the total amount spent on participant compensation?
    \answerNA{}
\end{enumerate}

\end{enumerate}

\newpage
\appendix
\section{Missing Proofs}
\newenvironment{assumption-multiplex}[1]
  {\renewcommand{\theassumption}{\ref{#1}$'$}%
   \addtocounter{assumption}{-1}%
   \begin{assumption}}
  {\end{assumption}}
In this section, we will proof the main theorem in our paper. Before the detailed proof, we first recall the following assumptions that are commonly used for characterizing the convergence of non-convex stochastic optimization.
\begin{assumption-multiplex}{assume:bounded-gradient}
(Bounded Gradient.) It exists $G \geq 0$ s.t. $||\nabla f(\w)|| \leq G$.
\end{assumption-multiplex}

\begin{assumption-multiplex}{assume:bounded-variance}
(Bounded Variance.) It exists $\sigma \geq 0$ s.t. $\E [||g(\w) - \nabla f(\w)||^2] \leq \sigma^2$.
\end{assumption-multiplex}

\begin{assumption-multiplex}{assume:l-smoothness}
(L-smoothness.) It exists $L > 0 $ s.t. $||\nabla f(\w) - \nabla f(\boldsymbol{v})|| \leq L||\w - \boldsymbol{v}|| $, $\forall  \w, \boldsymbol{v} \in \mathbb{R}^d$.
\end{assumption-multiplex}

\subsection{Proof of Theorem 1}
Based on the objective function of Sharpness-Aware Minimization~(SAM), suppose we can obtain the noisy observation gradient $g(\w)$ of true gradient $\nabla f(\w)$, we can write the iteration of SAM:
\begin{equation}
\label[equation]{equ:iter-sam}
\left\{
    \begin{array}{ll}
        \w_{t+\frac{1}{2}}=&\w_t + \rho \cdot \frac{g(\w_t)}{||g(\w_t)||} \\
        \w_{t+1}=&\w_t - \eta \cdot g(\w_{t+\frac{1}{2}})
    \end{array}
\right.
\end{equation}

\begin{lemma}
\label{lemma:1}
For any $\rho > 0$, $L > 0$ and the differentiable function $f$, we have the following inequality:
\begin{align*}
    \langle \nabla f(\w_{t}), \nabla f(\w_t + \rho \frac{\nabla f(\w_t)}{||\nabla f(\w_t)||})\rangle 
    \geq 
    ||\nabla f(\w_t)||^2 - \rho L G
\end{align*}
\end{lemma}
\begin{proof}
We first add and subtract a term $||\nabla f(\w_t)||$ to make use of classical inequalities bounding $\langle \nabla f(\w_1) - \nabla f(\w_2), \w_1 - \w_2 \rangle$ by $||\w_1 - \w_2||^2$ for smooth.

\begin{align*}
LHS
=& \langle \nabla f(\w_t) ,\nabla f(\w_t + \rho \frac{\nabla f(\w_t)}{||\nabla f(\w_t)||}) - \nabla f(\w_t) \rangle + ||\nabla f(\w_t)||^2
\\
=& \frac{||\nabla f(\w_t)||}{\rho} \langle \frac{\rho}{||\nabla f(\w_t)||}\nabla f(\w_t) ,\nabla f(\w_t + \rho \frac{\nabla f(\w_t)}{||\nabla f(\w_t)||})) - \nabla f(\w_t) \rangle + ||\nabla f(\w_t)||^2
\\
\geq&  -L \frac{||\nabla f(\w_t)||}{\rho} ||\frac{\rho}{||\nabla f(\w_t)||}\nabla f(\w_t)||^2 + ||\nabla f(\w_t)||^2
\\
=& - ||\nabla f(\w_t)|| \rho L + ||\nabla f(\w_t)||^2
\\
\geq& - G \rho L + ||\nabla f(\w_t)||^2
\end{align*}

where the first inequality is that
\begin{align*}
    \langle \nabla f(\w_1)-\nabla f(\w_2), \w_1 - \w_2 \rangle \geq -L||\w_1 - \w_2||^2,
\end{align*}
and the second inequality is the Assumption~\ref{assume:bounded-gradient}.
\end{proof}

\begin{lemma}
\label{lemma:2}
For $\rho > 0$, $L>0$, the iteration~\ref{equ:iter-sam} satisfies following inequality:
\begin{align*}
    \E\langle \nabla f(\w_t), g(\w_{t + \frac{1}{2}})\rangle 
    \geq
    \frac{1}{2}||\nabla f(\w_t)||^2 - L^2 \rho^2 - L \rho G
\end{align*}
\end{lemma}

\begin{proof}
We denote the deterministic values of $\w_{t+\frac{1}{2}}$ as $\hat{\w}_{t+\frac{1}{2}} = \w_t + \rho \frac{\nabla f(\w_t)}{||\nabla f(\w_t)||}$ in this section. After we add and subtract the term $g(\hat{\w}_{t+\frac{1}{2}})$, we have the following equation:
\begin{align*}
    \langle \nabla f(\w_t), g(\w_{t + \frac{1}{2}})\rangle 
    =&
    \langle \nabla f(\w_t), g(\w_t + \rho \frac{g(\w_t)}{||g(\w_t)||}) - g(\hat{\w}_{t+\frac{1}{2}})\rangle  + \langle g(\hat{\w}_{t+\frac{1}{2}}), \nabla f(\w_t) \rangle
\end{align*}
For the first term, we bound it by using the smoothness of $g(\w)$:
\begin{align*}
    - \langle \nabla f(\w_t), g(\w_t + \rho \frac{g(\w_t)}{||g(\w_t)||}) - g(\hat{\w}_{t+\frac{1}{2}})\rangle
    \leq& \frac{1}{2}||g(\w_t + \rho \frac{g(\w_t)}{||g(\w_t)||}) - g(\hat{\w}_{t+\frac{1}{2}})||^2 + \frac{1}{2}||\nabla f(\w_t)||^2
    \\
    \leq& \frac{L^2}{2} ||\w_t + \rho \frac{g(\w_t)}{||g(\w_t)||} - \hat{\w}_{t+\frac{1}{2}}||^2 + \frac{1}{2}||\nabla f(\w_t)||^2
    \\
    =& \frac{L^2}{2} ||\w_t + \rho \frac{g(\w_t)}{||g(\w_t)||} - (\w_t+\rho \frac{\nabla f(\w_t)}{||\nabla f(\w_t)||})||^2 + \frac{1}{2}||\nabla f(\w_t)||^2
    \\
    =& \frac{L^2\rho^2}{2} ||\frac{g(\w_t)}{||g(\w_t)||} - \frac{\nabla f(\w_t)}{||\nabla f(\w_t)||}||^2 + \frac{1}{2}||\nabla f(\w_t)||^2
    \\
    \leq& L^2 \rho^2 + \frac{1}{2}||\nabla f(\w_t)||^2
\end{align*}
For the second term, by using the Lemma~\ref{lemma:1}, we have:
\begin{align*}
    \E\langle g(\hat{\w}_{t+\frac{1}{2}}), \nabla f(\w_t) \rangle
    =&
    \langle \nabla f(\hat{\w}_{t+\frac{1}{2}}), \nabla f(\w_t) \rangle
    \\
    =& \langle \nabla f(\w_t + \rho \frac{\nabla f(\w_t)}{||\nabla f(\w_t)||}), \nabla f(\w_t) \rangle
    \\
    \geq& ||\nabla f(\w_t)||^2 - \rho L G
\end{align*}
Asesmbling the two inequalities yields to the result. 
\end{proof}

\begin{lemma}
\label{lemma:3}
For $\eta \leq \frac{1}{L}$, the iteration~\ref{equ:iter-sam} satisfies for all $t>0$:
\begin{align*}
    \E f(\w_{t+1}) \leq \E f(\w_t) - \frac{\eta}{2}\E||\nabla f(\w_t)||^2 + L\eta^2\sigma^2 +\eta L^2\rho^2 + (1-L\eta)\eta LG\rho
\end{align*}
\end{lemma}

\begin{proof}
By the smoothness of the function $f$, we obtain
\begin{align*}
    f(\w_{t+1}) 
    \leq& 
    f(\w_t) - \eta \langle \nabla f(\w_t), g(\w_{t+\frac{1}{2}}) \rangle + \frac{L\eta^2}{2}||g(\w_{t+\frac{1}{2}})||^2
    \\
    =& f(\w_t) - \eta \langle \nabla f(\w_t), g(\w_{t+\frac{1}{2}}) \rangle + \frac{L\eta^2}{2} ( ||\nabla f(\w_t)-g(\w_{t+\frac{1}{2}})||^2-||\nabla f(\w_t)||^2 + 2\langle \nabla f(\w_t), g(\w_{t+\frac{1}{2}})\rangle)
    \\
    =& f(\w_t) -\frac{L\eta^2}{2}||\nabla f(\w_t)||^2 + \frac{L\eta^2}{2} ||\nabla f(\w_t) - g(\w_{t+\frac{1}{2}})||^2 - (1-L\eta)\eta \langle  \nabla f(\w_t), g(\w_{t+\frac{1}{2}})\rangle
    \\
    \leq& f(\w_t) - \frac{L\eta^2}{2}||\nabla f(\w_t)||^2 + L\eta^2 ||\nabla f(\w_t) - g(\w_t)||^2 + L\eta^2 ||g(\w_t) - g(\w_{t+\frac{1}{2}})||^2 \\& - (1-L\eta)\eta \langle  \nabla f(\w_t), g(\w_{t+\frac{1}{2}})\rangle
    \\
    \leq& f(\w_t) - \frac{L\eta^2}{2}||\nabla f(\w_t)||^2 + L\eta^2 ||\nabla f(\w_t) - g(\w_t)||^2 + L\eta^2 L^2||\w_t - \w_{t+\frac{1}{2}}||^2 \\& - (1-L\eta)\eta \langle  \nabla f(\w_t), g(\w_{t+\frac{1}{2}})\rangle
    \\
    =& f(\w_t) - \frac{L\eta^2}{2}||\nabla f(\w_t)||^2 + L\eta^2 ||\nabla f(\w_t) - g(\w_t)||^2 + \eta^2L^3 \rho^2
    \\& -  (1-L\eta)\eta \langle  \nabla f(\w_t), g(\w_{t+\frac{1}{2}})\rangle
\end{align*}
Taking the expectation and using Lemma~\ref{lemma:2} we obtain
\begin{align*}
\E f(\w_{t+1}) 
\leq& 
\E f(\w_t) - \frac{L\eta^2}{2}\E ||\nabla f(\w_t)||^2 + L\eta^2\E ||\nabla f(\w_t) - g(\w_t)||^2 + \eta^2L^3 \rho^2 \\& - (1-L\eta)\eta \E \langle  \nabla f(\w_t), g(\w_{t+\frac{1}{2}})\rangle
\\
\leq& \E f(\w_t) - \frac{L\eta^2}{2}\E ||\nabla f(\w_t)||^2 + L\eta^2 \sigma^2 + \eta^2L^3 \rho^2 \\& - (1-L\eta)\eta \E \langle  \nabla f(\w_t), g(\w_{t+\frac{1}{2}})\rangle
\\
\leq& \E f(\w_t) - \frac{L\eta^2}{2}\E ||\nabla f(\w_t)||^2 + L\eta^2 \sigma^2 + \eta^2L^3 \rho^2 \\& - (1-L\eta)\eta \left[\frac{1}{2}\E||\nabla f(\w_t)||^2 - L^2\rho^2 -L\rho G\right]
\\
=& \E f(\w_t) - \frac{\eta}{2} \E ||\nabla f(\w_t)||^2 + L\eta^2 \sigma^2 + \eta^2L^3 \rho^2 + (1-L\eta)\eta  L^2\rho^2 + (1-L\eta)\eta L\rho G
\\
=& \E f(\w_t) - \frac{\eta}{2} \E ||\nabla f(\w_t)||^2 + L\eta^2 \sigma^2 + \eta L^2 \rho^2 + (1-L\eta)\eta L G \rho
\end{align*}
\end{proof}

\begin{proposition}
\label[Proposition]{pro:1}
Let $\eta_t = \frac{\eta_0}{\sqrt{t}}$ and perturbation amplitude $\rho$ decay with square root of $t$, \emph{e.g.}, $\rho_t=\frac{\rho_0}{\sqrt{t}}$. For $\rho_0 \leq G \eta_0$ and $\eta_0 \leq \frac{1}{L}$, we have

\begin{align*}
    \frac{1}{T} \sum_{t=1}^T \E ||\nabla f(\w_t)||^2 \leq& C_1 \frac{1}{\sqrt{T}} + C_2\frac{\log T}{\sqrt{T}},
\end{align*}
where $C_1=\frac{2}{\eta_0}(f(\w_0)-\E f(\w_T))$ and $C_2=2 (L \sigma^2 \eta_0 + LG\rho_0)$.

\begin{proof}
By Lemma~\ref{lemma:3}, we replace $\rho$ and $\eta$ with $\rho_t=\frac{\rho_0}{\sqrt{t}}$ and $\eta_t = \frac{\eta_0}{\sqrt{t}}$, we have
\begin{align*}
   \E f(\w_{t+1}) \leq& \E f(\w_t) - \frac{\eta_t}{2} \E ||\nabla f(\w_t)||^2 + L\eta_t^2 \sigma^2 + \eta_t L^2 \rho_t^2 + (1-L\eta_t)\eta_t L G \rho_t.
\end{align*}
Take telescope sum, we have
\begin{align*}
    \E f(\w_{T}) - f(\w_0) \leq&  - \sum_{t=1}^T \frac{\eta_t}{2} \E ||\nabla f(\w_t)||^2 + (L \sigma^2 \eta_0^2 + LG\rho_0\eta_0) \sum_{t=1}^T\frac{1}{t} + (L^2\eta_0\rho_0^2 - L^2G\eta_0^2\rho_0)\sum_{t=1}^T\frac{1}{t^{\frac{3}{2}}} 
\end{align*}
Under $\rho_0 \leq G \eta_0$, the last term will be less than 0, which means:
\begin{align*}
    \E f(\w_{T}) - f(\w_0) \leq&  - \sum_{t=1}^T \frac{\eta_t}{2} \E ||\nabla f(\w_t)||^2 + (L \sigma^2 \eta_0^2 + LG\rho_0\eta_0) \sum_{t=1}^T\frac{1}{t}.
\end{align*}
With 
\begin{align*}
    \frac{\eta_T}{2}\sum_{t=1}^T \E||\nabla f(\w_t)||^2 \leq& \sum_{t=1}^T \frac{\eta_t}{2} \E ||\nabla f(\w_t)||^2 \leq f(\w_0) - \E f(\w_{T}) + (L \sigma^2 \eta_0^2 + LG\rho_0\eta_0) \sum_{t=1}^T\frac{1}{t},
\end{align*}
we have 
\begin{align*}
    \frac{\eta_0}{2\sqrt{T}}\sum_{t=1}^T \E||\nabla f(\w_t)||^2 \leq& f(\w_0) - \E f(\w_{T}) + (L \sigma^2 \eta_0^2 + LG\rho_0\eta_0) \sum_{t=1}^T\frac{1}{t}
    \\
    \leq& f(\w_0) - \E f(\w_{T}) + (L \sigma^2 \eta_0^2 + LG\rho_0\eta_0) \log T.
\end{align*}
Finally, we achieve the result:
\begin{align*}
    \frac{1}{T}\sum_{t=1}^T\E||\nabla f(\w_t)||^2 \leq& \frac{2 \cdot (f(\w_0) - \E f(\w_{T}))}{\eta_0}\frac{1}{\sqrt{T}} + 2 (L \sigma^2 \eta_0 + LG\rho_0)\frac{\log T}{\sqrt{T}},
\end{align*}
which shows that SAM can converge at the rate of $O(\log T/\sqrt{T})$.
\end{proof}
\end{proposition}

\subsection{Proof of Theorem 2}
Suppose we can obtain the noisy observation gradient $g(\w_t)$ of true gradient $\nabla f(\w_t)$, and the mask $\m$, we can write the iteration of SAM:
Consider the iteration of Sparse SAM:
\begin{align}
\label{equ:iter-ssam}
\left\{
    \begin{array}{ll}
        \tilde{\w}_{t+\frac{1}{2}}=&\w_t + \rho \frac{g(\w_t)}{||g(\w_t)||} \odot \m_t \\
        \w_{t+1}=&\w_t - g(\tilde{\w}_{t+\frac{1}{2}})
    \end{array}
\right.
\end{align}
Let us denote the difference as $\w_{t+\frac{1}{2}} - \tilde{\w}_{t+\frac{1}{2}}=\boldsymbol{e}_t$.

\begin{lemma}
\label{lemma:4}
With $\rho > 0$, we have:
\begin{align*}
    \E\langle \nabla f(\w_t), g(\tilde{\w}_{t + \frac{1}{2}})\rangle 
    \geq
    \frac{1}{2}||\nabla f(\w_t)||^2 - 2 L^2 \rho^2 - L \rho G - L^2 ||\boldsymbol{e}_t||^2
\end{align*}

\begin{proof}
Similar to Lemma~\ref{lemma:2}, We denote the true gradient as $\hat{\w}_{t+\frac{1}{2}} = \w_t + \rho \frac{\nabla f(\w_t)}{||\nabla f(\w_t)||}$, and also add and subtract the item $g(\tilde{\w}_{t + \frac{1}{2}})$:
\begin{align*}
    \langle \nabla f(\w_t), g(\tilde{\w}_{t + \frac{1}{2}})\rangle 
    =&
    \langle \nabla f(\w_t), g(\tilde{\w}_{t + \frac{1}{2}}) - g(\hat{\w}_{t+\frac{1}{2}})\rangle  + \langle g(\hat{\w}_{t+\frac{1}{2}}), \nabla f(\w_t) \rangle
\end{align*}
For the first term, we bound it by using the smoothness of $g(\w)$:
\begin{align*}
    - \langle \nabla f(\w_t), g(\tilde{\w}_{t + \frac{1}{2}}) - g(\hat{\w}_{t+\frac{1}{2}})\rangle
    \leq& \frac{1}{2}||g(\tilde{\w}_{t + \frac{1}{2}}) - g(\hat{\w}_{t+\frac{1}{2}})||^2 + \frac{1}{2}||\nabla f(\w_t)||^2
    \\
    \leq& \frac{L^2}{2} ||\tilde{\w}_{t + \frac{1}{2}} - \hat{\w}_{t+\frac{1}{2}}||^2 + \frac{1}{2}||\nabla f(\w_t)||^2
    \\
    =& \frac{L^2}{2} ||\w_{t + \frac{1}{2}} - \boldsymbol{e}_t - \hat{\w}_{t+\frac{1}{2}}||^2 + \frac{1}{2}||\nabla f(\w_t)||^2
    \\
    \leq& L^2 (||\w_{t+\frac{1}{2}} - \hat{\w}_{t+\frac{1}{2}}||^2 + ||\boldsymbol{e}_t||^2) + \frac{1}{2}||\nabla f(\w_t)||^2
    \\
    =& L^2 (\rho^2 ||\frac{g(\w_t)}{||g(\w_t)||} - \frac{\nabla f(\w_t)}{||\nabla f(\w_t)||}||^2 + ||\boldsymbol{e}_t||^2) + \frac{1}{2}||\nabla f(\w_t)||^2
    \\
    \leq& 2L^2 \rho^2 + L^2||\boldsymbol{e}_t||^2 + \frac{1}{2}||\nabla f(\w_t)||^2
\end{align*}
For the second term, we do the same in Lemma~\ref{lemma:2}:
\begin{align*}
    \E\langle g(\hat{\w}_{t+\frac{1}{2}}), \nabla f(\w_t) \rangle
    \geq& ||\nabla f(\w_t)||^2 - \rho L G.
\end{align*}
Assembling the two inequalities yields to the result. 
\end{proof}
\end{lemma}

\begin{lemma}
For $\eta \leq \frac{1}{L}$, the iteration~\ref{equ:iter-ssam} satisfies for all $t>0$:
\begin{align*}
    \E f(\w_{t+1}) \leq& \E f(\w_t) - \frac{\eta}{2} \E ||\nabla f(\w_t)||^2 + L\eta^2\sigma^2 + 2\eta L^2\rho^2 + (1-L\eta)\eta LG\rho \\
    & + (1+L\eta)\eta L^2||\boldsymbol{e}_t||^2
\end{align*}
\end{lemma}

\begin{proof}
By the smoothness of the function $f$, we obtain
\begin{align*}
    f(\w_{t+1}) 
    \leq& 
    f(\w_t) - \eta \langle \nabla f(\w_t), g(\tilde{\w}_{t+\frac{1}{2}}) \rangle + \frac{L\eta^2}{2}||g(\tilde{\w}_{t+\frac{1}{2}})||^2
    \\
    =& f(\w_t) - \eta \langle \nabla f(\w_t), g(\tilde{\w}_{t+\frac{1}{2}}) \rangle + \frac{L\eta^2}{2} ( ||\nabla f(\w_t)-g(\tilde{\w}_{t+\frac{1}{2}})||^2-||\nabla f(\w_t)||^2 + 2\langle \nabla f(\w_t), g(\tilde{\w}_{t+\frac{1}{2}})\rangle)
    \\
    =& f(\w_t) -\frac{L\eta^2}{2}||\nabla f(\w_t)||^2 + \frac{L\eta^2}{2} ||\nabla f(\w_t) - g(\tilde{\w}_{t+\frac{1}{2}})||^2 - (1-L\eta)\eta \langle  \nabla f(\w_t), g(\tilde{\w}_{t+\frac{1}{2}})\rangle
    \\
    \leq& f(\w_t) - \frac{L\eta^2}{2}||\nabla f(\w_t)||^2 + L\eta^2 ||\nabla f(\w_t) - g(\w_t)||^2 + L\eta^2 ||g(\w_t) - g(\tilde{\w}_{t+\frac{1}{2}})||^2 \\& - (1-L\eta)\eta \langle  \nabla f(\w_t), g(\tilde{\w}_{t+\frac{1}{2}})\rangle
    \\
    \leq& f(\w_t) - \frac{L\eta^2}{2}||\nabla f(\w_t)||^2 + L\eta^2 ||\nabla f(\w_t) - g(\w_t)||^2 + L\eta^2 L^2||\w_t - \tilde{\w}_{t+\frac{1}{2}}||^2 \\& - (1-L\eta)\eta \langle  \nabla f(\w_t), g(\tilde{\w}_{t+\frac{1}{2}})\rangle
    \\
    =& f(\w_t) - \frac{L\eta^2}{2}||\nabla f(\w_t)||^2 + L\eta^2 ||\nabla f(\w_t) - g(\w_t)||^2 + \eta^2L^3 ||\w_t - \w_{t+\frac{1}{2}} + \boldsymbol{e}_t||^2
    \\& -  (1-L\eta)\eta \langle  \nabla f(\w_t), g(\tilde{\w}_{t+\frac{1}{2}} )\rangle
    \\
    \leq& f(\w_t) - \frac{L\eta^2}{2}||\nabla f(\w_t)||^2 + L\eta^2 ||\nabla f(\w_t) - g(\w_t)||^2 + 2 \eta^2L^3 (||\w_t - \w_{t+\frac{1}{2}}||^2 + ||\boldsymbol{e}_t||^2)
    \\& -  (1-L\eta)\eta \langle  \nabla f(\w_t), g(\tilde{\w}_{t+\frac{1}{2}} )\rangle
    \\
    =& f(\w_t) - \frac{L\eta^2}{2}||\nabla f(\w_t)||^2 + L\eta^2 ||\nabla f(\w_t) - g(\w_t)||^2 + 2 \eta^2L^3 ( \rho^2 + ||\boldsymbol{e}_t||^2)
    \\& -  (1-L\eta)\eta \langle  \nabla f(\w_t), g(\tilde{\w}_{t+\frac{1}{2}} )\rangle
\end{align*}
Taking the expectation and using Lemma~\ref{lemma:4} we obtain
\begin{align*}
\E f(\w_{t+1}) 
\leq& 
\E f(\w_t) - \frac{L\eta^2}{2}\E ||\nabla f(\w_t)||^2 + L\eta^2\E ||\nabla f(\w_t) - g(\w_t)||^2 + 2\eta^2L^3 ( \rho^2 + ||\boldsymbol{e}_t||^2) \\& - (1-L\eta)\eta \E \langle  \nabla f(\w_t), g(\w_{t+\frac{1}{2}})\rangle
\\
\leq& \E f(\w_t) - \frac{L\eta^2}{2}\E ||\nabla f(\w_t)||^2 + L\eta^2 \sigma^2 + 2\eta^2L^3 ( \rho^2 + ||\boldsymbol{e}_t||^2) \\& - (1-L\eta)\eta \E \langle  \nabla f(\w_t), g(\w_{t+\frac{1}{2}})\rangle
\\
\leq& \E f(\w_t) - \frac{L\eta^2}{2}\E ||\nabla f(\w_t)||^2 + L\eta^2 \sigma^2 + 2\eta^2L^3 ( \rho^2 + ||\boldsymbol{e}_t||^2) \\& - (1-L\eta)\eta \left[\frac{1}{2}||\nabla f(\w_t)||^2 - 2 L^2 \rho^2 - L \rho G - L^2 ||\boldsymbol{e}_t||^2\right]
\\
=& \E f(\w_t) - \frac{\eta}{2} \E ||\nabla f(\w_t)||^2 + L\eta^2\sigma^2 + 2\eta L^2\rho^2 + (1-L\eta)\eta LG\rho \\& + (1+L\eta)\eta L^2||\rho \frac{g(\w_t)}{||g(\w_t)||} \odot \m_t - \rho \frac{g(\w_t)}{||g(\w_t)||}||^2
\\
=& \E f(\w_t) - \frac{\eta}{2} \E ||\nabla f(\w_t)||^2 + L\eta^2\sigma^2 + 2\eta L^2\rho^2 + (1-L\eta)\eta LG\rho \\& + (1+L\eta)\eta L^2||\boldsymbol{e}_t||^2
\end{align*}
\end{proof}

\begin{proposition}
\label[Proposition]{pro:2}
Let us $\eta_t=\frac{\eta_0}{\sqrt{t}}$ and perturbation amplitude $\rho$ decay with square root of $t$, \emph{e.g.}, $\rho_t=\frac{\rho_0}{\sqrt{t}}$. With $\rho_0 \leq G\eta_0 / 2$, we have:
\begin{align*}
    \frac{1}{T}\sum_{t=1}^T \E||\nabla f(\w)||^2 \leq&
    C_3\frac{1}{\sqrt{T}} + C_4\frac{\log T}{\sqrt{T}},
\end{align*}
where $C_3=\frac{2}{\eta_0}(f(\w_0-\E f(\w_T)+\eta_0L^2\rho^2(1+\eta_0L)\frac{\pi^2}{6})$ and $C_4=2(L\sigma^2\eta_0+LG\rho_0)$.
\end{proposition}

\begin{proof}
By taking the expectation and using Lemma~\ref{lemma:4}, and taking the schedule to be $\eta_t = \frac{\eta_0}{\sqrt{t}}$, $\rho_t = \frac{\rho _0}{\sqrt{t}}$, we obtain:
\begin{align*}
    \E f(\w_{t+1}) \leq& \E f(\w_t) - \frac{\eta_t}{2} \E ||\nabla f(\w_t)||^2 + L\eta_t^2\sigma^2 + 2\eta_t L^2\rho_t^2 + (1-L\eta_t)\eta_t LG\rho_t \\& + (1+L\eta_t)\eta_t L^2||\boldsymbol{e}_t||^2
\end{align*}
By taking sum and bound $\rho$ with $\frac{G\eta_0}{2}$, we have:
\begin{align*}
    \frac{\eta_0}{2\sqrt{T}}\sum_{t=1}^T \E||\nabla f(\w_t)||^2 \leq& f(\w_0) - \E f(\w_{T}) + (L \sigma^2 \eta_0^2 + LG\rho_0\eta_0) \sum_{t=1}^T\frac{1}{t} \\ & + \sum_{t=1}^T(1+L\eta_t)\eta_t L^2||\boldsymbol{e}_t||^2
    \\
    \leq& f(\w_0) - \E f(\w_{T}) + (L \sigma^2 \eta_0^2 + LG\rho_0\eta_0) \sum_{t=1}^T\frac{1}{t} \\
    & + \eta_0 L^2 \rho_0^2 \sum_{t=1}^T\frac{1}{t^{\frac{3}{2}}} + \eta_0^2 L^3 \rho^2 \sum_{t=1}^T \frac{1}{t^2}
    \\
    \leq& f(\w_0) - \E f(\w_{T}) + (L \sigma^2 \eta_0^2 + LG\rho_0\eta_0) \sum_{t=1}^T\frac{1}{t} \\
    & +  \eta_0 L^2 \rho_0^2 \sum_{t=1}^T\frac{1}{t^2} + \eta_0^2 L^3 \rho_0^2 \sum_{t=1}^T \frac{1}{t^2}
    \\
    \leq& f(\w_0) - \E f(\w_{T}) + (L \sigma^2 \eta_0^2 + LG\rho_0\eta_0) \log T \\
    & + \eta_0L^2\rho_0^2(1+\eta_0L)\frac{\pi^2}{6}
    \\
\end{align*}
Finally, we achieve the result:
\begin{align*}
    \frac{1}{T}\sum_{t=1}^T\E ||\nabla f(\w_t)||^2\leq& \frac{2(f(\w_0-\E f(\w_T)+\eta_0L^2\rho^2(1+\eta_0L)\frac{\pi^2}{6})}{\eta_0}\frac{1}{\sqrt{T}} \\
    & + 2(L\sigma^2\eta_0+LG\rho_0)\frac{\log T}{\sqrt{T}}
\end{align*}
\end{proof}
So far, we have completed the proof of the theory in the main text.

\newpage

\section{More Experimets}

\textbf{VGG on CIFAR10.} To further confirm the model-agnostic characteristic of our Sparse SAM, we test the VGG-style architecture on CIFAR10. Following~\cite{pytorch-cifar100-github}, we test SSAM training the VGG11-BN on CIFAR10 and the results are shown in the following~\cref{table:vgg-cifar-ssam}. The perturbation magnitude $\rho$ is set to 0.05.

\begin{table}[ht]
\centering
\vspace{-2mm}
\caption{Test accuracy of VGG11-BN on CIFAR10 with proposed Sparse SAM.}
\label[table]{table:vgg-cifar-ssam}
\begin{tabular}{ccccc}
\hline
Model & Dataset & Optimizer & Sparsity & Accuracy \\ \hline
\multirow{8}{*}{VGG11-BN} & \multirow{8}{*}{CIFAR10} & SGD & / & 93.42\% \\ \cline{3-5} 
 &  & SAM & 0\% & 93.87\% \\ \cline{3-5} 
 &  & \multirow{6}{*}{SSAM-F/SSAM-D} & 50\% & \textbf{94.03\%}/93.79\% \\
 &  &  & 80\% & 93.83\%/\textbf{93.95\%} \\
 &  &  & 90\% & 93.76\%/93.85\% \\
 &  &  & 95\% & 93.77\%/93.48\% \\
 &  &  & 98\% & 93.54\%/93.54\% \\
 &  &  & 99\% & 93.47\%/93.33\% \\ \hline
\end{tabular}%
\end{table}

\textbf{SAM with different perturbation magnitude $\rho$.}
We determine the perturbation magnitude $\rho$ by using grid search. We choose $\rho$ from the set $\{0.01, 0.02, 0.05, 0.1, 0.2, 0.5\}$ for CIFAR, and choose $\rho$ from $\{0.01, 0.02, 0.05, 0.07, 0.1, 0.2\}$ for ImageNet. We show the results when varying $\rho$ in~\cref{table:rho-cifar} and~\cref{table:rho-imagenet}. From this table, we can see that the $\rho=0.1$, $\rho=0.2$ and $\rho=0.07$ is sutiable for CIFAR10, CIFAR100 and ImageNet respectively.

\begin{table}[ht]
\centering
\vspace{-2mm}
\caption{Test accuracy of ResNet18 and WideResNet28-10 on CIFAR10 and CIFAR100 with different perturbation magnitude $\rho$.}
\label[table]{table:rho-cifar}
\resizebox{1\textwidth}{12mm}{
\begin{tabular}{lccccccc}
\toprule
Dataset & SAM $\rho$ & 0.01    & 0.02    & 0.05    & 0.1              & 0.2              & 0.5     \\ \hline
         & ResNet18   & 96.58\% & 96.54\% & 96.68\% & \textbf{96.83\%} & 96.32\%          & 93.16\% \\
\multirow{-2}{*}{CIFAR10}  & WideResNet28-10 & 97.26\% & 97.34\% & 97.31\% & \textbf{97.48\%} & 97.29\%          & 95.13\% \\ \hline
         & ResNet18   & 79.56\% & 79.98\% & 80.71\% & 80.65\% & \textbf{81.03\%} & 77.57\% \\
\multirow{-2}{*}{CIFAR100} & WideResNet28-10 & 82.25\% & 83.04\% & 83.47\% & 83.47\% & \textbf{84.20\%} & 84.03\% \\ \bottomrule
\end{tabular}
}
\end{table}

\begin{table}[ht]
\centering
\vspace{-3mm}
\caption{Test accuracy of ResNet50 on ImageNet with different perturbation magnitude $\rho$.}

\label{table:rho-imagenet}
\begin{tabular}{cccccccc}
\toprule
datasets & SAM $\rho$ & 0.01    & 0.02    & 0.05    & 0.07             & 0.1     & 0.2     \\ \hline
ImageNet & ResNet50   & 76.63\% & 76.78\% & 77.12\% & \textbf{77.25\%} & 77.00\% & 76.37\% \\ \bottomrule
\end{tabular}
\end{table}

\textbf{Ablations of Masking Strategy.} For further verification of our masking strategy, we perform more ablations in this paragraph. For the mask update in SSAM-F, the parameters with largest fisher information are selected. Compared with SSAM-F, we consider the random mask,~\emph{i.e.}, the mask is randomly generated to choose which parameters are perturbated. For  the mask update in SSAM-D, we first drop the flattest weights and then random grow some weights. Compared with  SSAM-D, we experiment the SSAM-D which drops randomly or drops the sharpest weights,~\emph{i.e.}, the weights with large gradients. The results of ablations are shown in~\cref{table:mask-strategy}. The results show that random strategies are less effective than our SSAM. The performance of SSAM-D dropping sharpest weights drops a lot even worse than random strategy, which is consistent with our conjecture.

\begin{table}[h!]
\centering
\caption{Ablation of different masking strategy.}
\label[table]{table:mask-strategy}
\begin{tabular}{ccccc}
\toprule
Model & Dataset & Optimizer & Strategy & Accuracy \\ \hline
\multirow{7}{*}{ResNet50} & \multirow{7}{*}{ImageNet} & SGD & / & 76.67\% \\
 &  & SAM & / & 77.25\% \\ \cline{3-5} 
 &  & Sparse SAM & Random Mask & 77.08\%\textcolor{blue}{(-0.17)} \\
 &  & SSAM-F & Topk Fisher Information & 77.31\%\textcolor{blue}{(+0.06)} \\ \cline{3-5} 
 &  & \multirow{3}{*}{SSAM-D} & Random Drop & 77.08\%\textcolor{blue}{(-0.17)} \\
 &  &  & Drop Sharpnest weights & 76.68\%\textcolor{blue}{(-0.57)} \\
 &  &  & Drop Flattest weights & 77.25\%\textcolor{blue}{(-0.00)} \\ \bottomrule
\end{tabular}
\end{table}

\textbf{Influence of hyper-parameters}
We first examine the effect of the number of sample size $N_F$ of SSAM-F in~\cref{ablation-num-samples}. From it we can see that a certain number of samples is enough for the approximation of data distribution in SSAM-F,~\emph{e.g.}, $N_F=128$, which greatly saves the computational cost of SSAM-F. In~\cref{ablation-update-mask}, we also report the influence of the mask update interval on SSAM-F and SSAM-D. The results show that the performance degrades as the interval becom longer, suggesting that dense mask updates are necessary for our methods. Both of them are ResNet18 on CIFAR10.

\begin{table}[ht]
	\begin{minipage}[t]{0.53\textwidth}
	\small
		\centering
		\caption{Results of ResNet18 on CIFAR10 with different number of samples $N_F$ in SSAM-F. `Time' reported in table is the time cost to calculate Fisher Information based on $N_F$ samples.}
		\label{ablation-num-samples}
		\begin{tabular}{cccc}
        \toprule
        Sparsity & $N_F$ & Acc & Time \\ \hline
        \multirow{6}{*}{0.5} & 16 & 96.77\% & 1.49s \\ 
         & 128 & 96.84\% & 4.40s \\ 
         & 512 & 96.67\% & 15.35s \\ 
         & 1024 & 96.83\% & 30.99s \\ 
         & 2048 & 96.68\% & 56.23s \\ 
         & 4096 & 96.66\% & 109.31s \\ \hline
         \multirow{6}{*}{0.9} & 16 &  96.79\% & 1.47s\\
         & 128 & 96.50\% & 5.42s\\
         & 512 & 96.43\% & 15.57s\\
         & 1024 & 96.75\% & 29.24s \\
         & 2048 & 96.62\% & 57.72s\\
         & 4096 & 96.59\% & 110.65s\\ \bottomrule
        \end{tabular}
	\end{minipage}
	\hspace{+3mm}
	\begin{minipage}[t]{0.43\textwidth}
	\small
	\caption{Results of ResNet18 on CIFAR10 with different $T_m$ intervals of update mask. The left of `/' is accuracy of SSAM-F, while the right is SSAM-D.}
		\label{ablation-update-mask}
		\centering
		\begin{tabular}{ccc}
        \toprule
        Sparsity & $T_m$ & Acc \\ \hline
        \multirow{6}{*}{0.5} & 1 & 96.81\%/96.74\% \\
         & 2 & 96.51\%/96.74\% \\
         & 5 & 96.83\%/96.60\% \\
         & 10 & 96.71\%/96.73\% \\
         & 50 & 96.65\%/96.75\% \\
         & Fixed & 96.57\%/96.52\% \\ \hline
        \multirow{6}{*}{0.9} & 1 & 96.70\%/96.65\% \\
         & 2 & 93.75\%/96.63\% \\
         & 5 & 96.51\%/96.69\% \\
         & 10 & 96.67\%/96.74\% \\
         & 50 & 96.64\%/96.66\% \\
         & Fixed & 96.21\%/96.46\% \\ \bottomrule
        \end{tabular}
	\end{minipage}
\end{table}

\section{Limitation and Societal Impacts}

\textbf{Limitation.} Our method Sparse SAM is mainly based on sparse operation. At present, the sparse operation that has been implemented is only 2:4 sparse operation. The 2:4 sparse operation requires that there are at most two non-zero values in four contiguous memory, which does not hold for us. To sum up, there is currently no concrete implemented sparse operation to achieve training acceleration. But in the future, with the development of hardware for sparse operation, our method has great potential to achieve truly training  acceleration.

\textbf{Societal Impacts.} 
In this paper, we provide a Sparse SAM algorithm that reduces computation burden and improves model generalization. In the future, we believe that with the development of deep learning, more and more models need the guarantee of generalization and also the efficient training. Different from the work on sparse networks, our proposed Sparse SAM does not compress the model for hardware limited device, but instead accelerates model training. It's helpful for individuals or laboratories which are lack computing resources.


\end{document}